\documentclass[journal]{IEEEtran}
\usepackage{amsmath,amsfonts}
\usepackage{array}
\usepackage{algorithmic}
\usepackage[ruled,lined,linesnumbered,ruled]{algorithm2e}  
\usepackage{multirow}
\usepackage{graphicx} 
\usepackage{float} 
\usepackage[caption=false,font=footnotesize,labelfont=rm,textfont=rm]{subfig}
\usepackage{CJKutf8}
\usepackage{textcomp}
\usepackage{stfloats}
\usepackage{url}
\usepackage{verbatim}
\usepackage{graphicx}
\usepackage{cite}
\begin{document}
\begin{CJK}{UTF8}{gbsn}
\title{\textbf{Recurrent Stochastic Configuration Networks for Temporal Data Analytics}}
\author{
  Dianhui Wang 
  \thanks{\textit{\underline{Corresponding author}}: 
\textbf{dh.wang@deepscn.com}}\\
  State Key Laboratory of Synthetical Automation for Process Industries \\
  Northeastern University, Shenyang 110819, China \\
 Research Center for Stochastic Configuration Machines\\
  China University of Mining and Technology, Xuzhou 221116, China\\
  Gang Dang \\
  State Key Laboratory of Synthetical Automation for Process Industries \\
  Northeastern University, Shenyang 110819, China\\
}
\maketitle
\newtheorem{remark}{\bf Remark}

\begin{abstract}
Temporal data modelling techniques with neural networks are useful in many domain applications. This paper aims at developing a recurrent version of stochastic configuration networks (RSCNs) for problem solving, where we have no underlying assumption on the dynamic orders of the input variables. Given a collection of historical data, we first build an initial RSCN model in the light of a supervisory mechanism, followed by an online update of the output weights using projection algorithms. Some theoretical results are established, including the echo state property, the universal approximation property of RSCN for the offline learning, the convergence of the output weights of RSCN for the online learning, and the stability of the modelling performance. The proposed RSCN model is remarkably distinguished from the well-known echo state networks (ESNs) in terms of the way of assigning the input random weight matrix and a special structure of the random feedback matrix. A comprehensive comparison study among the long short-term memory (LSTM) network, the original ESN, and several state-of-the-art ESN methods such as the simple cycle reservoir (SCR), the polynomial ESN (PESN), the leaky-integrator ESN (LIESN) and RSCN is carried out. Experimental results clearly indicate that the proposed RSCN performs favourably over all of the testing datasets.
\end{abstract}

\begin{IEEEkeywords}
Recurrent stochastic configuration networks, echo state property, universal approximation property, temporal data analytics.
\end{IEEEkeywords}

\section{Introduction}
\IEEEPARstart{N}{owdays}, temporal data is prevalent across various industrial processes, where analyzing and utilizing this information effectively is crucial for informed decision-making, resource allocation optimization, and future trend forecasting \cite{ref1,ref3}. However, successfully modelling uncertain dynamics still remains a challenge due to their complex input-output dependencies and nonlinear relationships. Neural networks (NNs), particularly recurrent neural networks (RNNs) and their variants, such as long short-term memory (LSTM) networks, have demonstrated great potential for problem solving \cite{ref4,ref5,ref6}. These models can capture the temporal characteristics within the data through their unique recurrent structures, making them well-suited for temporal data analytics. Unfortunately, training RNNs and LSTM networks based on the back-propagation (BP) algorithm suffers from the drawbacks of slow learning, sensitivity to the initialization of the weights and biases, and local minima. Therefore, it is essential to develop advanced temporal data modelling techniques that align with the industrial demands on response speed, stability, and forecasting accuracy.

In recent years, randomized neural networks have received considerable attention for temporal data analytics \cite{ref7,ref8,ref9}. Among them, echo state networks (ESNs) are a typical class of randomized models, which have been applied in various fields due to their effectiveness and efficiency \cite{ref10,ref11,ref12,ref13}. The original ESN can be regarded as a three-layer neural network comprising an input layer with fixed connections, a sparsely connected reservoir, and a readout layer. Only the output weights need to be calculated using the least squares method, effectively overcoming the drawbacks of the BP-based methods. Although ESNs present several advantages over traditional RNNs, they lack a basic understanding on both the structure design and the random parameters assignment. With such a background, parameter optimization strategies for improving the learning and predictive abilities of ESNs have been reported in \cite{ref14,ref15,ref16,ref17,ref18}. Specifically, it was found that the scale of random weights significantly influences the prediction accuracy. Nonetheless, the optimization process obviously increases the computational cost. To further enhance the prediction accuracy, researchers have concentrated on the design of reservoir topologies. Notable advancements include the simple cycle reservoir (SCR) for reducing the computational complexity of reservoir construction, growing ESNs (GESNs) for automatically designing the reservoir topology, leaky-integrator ESNs (LIESNs) for improving network flexibility, and deep ESNs for promoting the network representation capabilities, as discussed in \cite{ref19,ref20,ref21,ref22}. However, these approaches cannot guarantee the model's universal approximation property, which is fundamental for data modelling theory and practice. Our findings reported in \cite{ref23} substantially improve the existing randomized learning techniques with deeper insights into the way of assigning the random parameters. Distinguished from the works presented in \cite{ref231,ref232,ref24,ref25}, stochastic configuration networks (SCNs) are incrementally constructed in the light of a supervisory mechanism, resulting in a class of randomized universal approximators.

Built on the SCN concept, in this paper we develop a recurrent version of SCNs (RSCNs) for temporal data analytics. The RSCN can be regarded as an incrementally built ESN that inherits the echo state property (ESP) while addressing the issues of parameter selection and structure setting. Given a collection of historical data, the initial model can be trained using our proposed recurrent stochastic configuration (RSC) algorithm. Then, the output weights are dynamically adjusted according to some projection algorithms \cite{ref35}. The proposed RSC learning algorithm ensures modelling performance for both the offline and online learnings. Experimental results demonstrate that RSCN outperforms other classical models in terms of learning and generalization performance, indicating its power in modelling complex temporal data. Our contributions of this paper can be summarized as follows.

\begin{itemize}
  \item [1)] 
  A novel design of randomized RNNs is proposed for temporal data modelling. This method utilizes the reservoir to process dynamic temporal information and generates random parameters in the light of a supervisory mechanism, effectively capturing temporal dependencies within the data while addressing the sensitivity of learning parameters.     
  \item [2)]
  A special structure of the random feedback matrix is defined and its spectral radius is constrained, which guarantees the universal approximation property and echo state property of the built model, resulting in superior nonlinear processing capability and stable output response. 
  \item [3)]
  Theoretical analyses on the convergence of the output weights, the stability of the modelling performance, the universal approximation property of the offline learning, and the echo state property are given.
  \item [4)]
  A set of comparisons are carried out, demonstrating the effectiveness of the proposed RSCN model and correctness of the theoretical results.
\end{itemize}

The remainder of this paper is organized as follows. Section II formulates the problem and reviews some related works. Section III details the proposed RSCN and learning algorithms. Section IV presents some theoretical results. Section V reports our experimental results with comparisons. Finally, Section VI concludes this paper.

\section{Related work}
In this section, some related work are introduced, including the problem formulation, the well-known echo state networks, and stochastic configuration networks.
\subsection{Problem formulation}
Consider a discrete unknown nonlinear system
\begin{equation}
\label{eq1}
\begin{array}{l}
y\left( {n + 1} \right) = {f_{\rm{p}}}\left( {y\left( n \right), \ldots ,y\left( {n - {d_A} + 1} \right),u\left( n \right),} \right.\\
{\kern 1pt} {\kern 1pt} {\kern 1pt} {\kern 1pt} {\kern 1pt} {\kern 1pt} {\kern 1pt} {\kern 1pt} {\kern 1pt} {\kern 1pt} {\kern 1pt} {\kern 1pt} {\kern 1pt} {\kern 1pt} {\kern 1pt} {\kern 1pt} {\kern 1pt} {\kern 1pt} {\kern 1pt} {\kern 1pt} {\kern 1pt} {\kern 1pt} {\kern 1pt} {\kern 1pt} {\kern 1pt} {\kern 1pt} {\kern 1pt} {\kern 1pt} {\kern 1pt} {\kern 1pt} {\kern 1pt} {\kern 1pt} {\kern 1pt} {\kern 1pt} {\kern 1pt} {\kern 1pt} {\kern 1pt} {\kern 1pt} {\kern 1pt} {\kern 1pt} {\kern 1pt} {\kern 1pt} {\kern 1pt} {\kern 1pt} {\kern 1pt} {\kern 1pt} {\kern 1pt} {\kern 1pt} {\kern 1pt} {\kern 1pt} \left. { \ldots ,u\left( {n - {d_B} + 1} \right)} \right) + e\left( n \right),
\end{array}
\end{equation}
where $y,u,e$ are the plant output, control input and process error, respectively, ${{d}_{A}}$ and ${{d}_{B}}$ are the dynamic orders for $y$ and $u$, ${{f}_{\text{p}}}$ is a nonlinear function. Due to external disturbances and changes in the process environment, time-varying delays and unknown orders may occur in the system, leading to poor modelling performance \cite{ref26,ref27}. Notice that the orders identification will inevitably increase the computational overhead. This paper aims to tackle the challenge of modelling nonlinear systems with uncertain dynamic orders. The output $y\left( n+1 \right)$ can be determined solely through $u( {n - {{\hat d}_B}} )$ and $y( {n - {{\hat d}_A}} )$, where ${\hat d}_B$ and ${\hat d}_A$ represent the dynamic orders derived from some prior knowledge.

Temporal data analytics primarily focuses on processing, analyzing, and interpreting time-related data to extract meaningful insights. This field encompasses not only time-series forecasting but also the analysis of dynamic behaviors in complex control systems. Time-series forecasting involves modelling the time dependence of an output variable $y$ and predicting future trends and fluctuations. Meanwhile, control system analysis explores the dynamic relationship between input $u$ and output $y$, using tools like transfer functions and state space models. In some scenarios, the input of a control system may not include $y$, offering greater flexibility in controller design. By selecting appropriate modelling and analysis approaches tailored to specific operating conditions, valuable patterns and insights can be effectively extracted from temporal data, offering robust support for practical applications.

In industrial processes, the information on the dynamic orders plays a crucial role in describing the relationship between the system output and inputs, particularly by accounting for time delays in the input signals \cite{ref28,ref29}. However, accurately determining the dynamic orders in real-world applications poses significant challenges due to the inherent complexity and non-stationary characteristics of industrial systems. While it may be feasible to estimate the dynamic order based on prior knowledge or experience, achieving the requisite level of accuracy for effective modelling often proves elusive. Traditional models may struggle to adequately capture these time-varying characteristics, resulting in suboptimal performance in predicting or controlling the system's behavior.

\subsection{Echo state networks}
ESNs exploit the reservoir to map the input signal to a high-dimensional and intricate state space. Unlike RNNs, ESNs randomly assign the input weights and biases, with output weights determined using the least squares method instead of iterative optimization. This helps to avoid the drawbacks of slow convergence and easily falling into local minima. Moreover, ESNs have the echo state property  \cite{ref10}, where the reservoir state $\mathbf{x}\left( n \right)$ is the echo of the input and $\mathbf{x}\left( n \right)$ should asymptotically depend on the driving input signal. Recent research on ESNs has focused on investigating the dynamic properties of the reservoir \cite{ref30}, parameters optimization \cite{ref15,ref16,ref31}, reservoir topology design \cite{ref32,ref33,ref34}, and their practical applications \cite{ref11,ref12,ref13}.

Consider an ESN model with $N$ reservoir nodes:
\begin{equation}
\label{eq2}
{\bf{x}}(n) = g({{\bf{W}}_{{\rm{in,}}N}}{\bf{u}}(n) + {{\bf{W}}_{{\mathop{\rm r}\nolimits} {\rm{,}}N}}{\bf{x}}(n - 1) + {{\bf{b}}_N}),
\end{equation}
\begin{equation}
\label{eq3}
{\bf{y}}(n) = {{\bf{W}}_{{\mathop{\rm out}\nolimits} {\rm{,}}N}}\left( {{\bf{x}}(n),{\bf{u}}(n)} \right),
\end{equation}
where $\mathbf{u}(n)\in {{\mathbb{R}}^{K}}$ is the input signal, $\mathbf{x}(n)\in {{\mathbb{R}}^{N}}$ is the internal state vector of the reservoir, ${{\mathbf{b}}_{N}}$ is the bias, ${{\bf{W}}_{{\mathop{\rm out}\nolimits} {\rm{,}}N}}\in {{\mathbb{R}}^{L\times \left( N+K \right)}}$ is the output weight matrix, $K$ and $L$ are the dimensions of input and output, respectively, and $g\left( x \right)$ is the activation function. The input weight matrix ${{\mathbf{W}}_{\operatorname{in},N}}$ and feedback matrix ${{\bf{W}}_{{\mathop{\rm r}\nolimits} {\rm{,}}N}}$ are generated from a uniform distribution over $\left[ -\lambda ,\lambda  \right]$, and they remain fixed once initialized. Define $\mathbf{X}\text{=}\left[ \left( \mathbf{x}(1),\mathbf{u}(1) \right),\ldots ,\left( \mathbf{x}({{n}_{\max }}),\mathbf{u}({{n}_{\max }}) \right) \right]$, where ${{n}_{\max }}$ is the number of samples, the model output is
\begin{equation}
\label{eq5}
{\bf{Y}} = \left[ {{\bf{y}}\left( 1 \right),{\bf{y}}\left( 2 \right),...,{\bf{y}}\left( {{n_{max}}} \right)} \right] = {\bf{W}}_{{\mathop{\rm out}\nolimits} ,N}^{}{\bf{X}}.
\end{equation}
The output weight matrix ${{\mathbf{W}}_{\rm out}}$ can be obtained by the least square method, that is,
\begin{equation}
\label{eq7}
{\bf{W}}_{{\mathop{\rm out}\nolimits} ,N}^ \top  = {\left( {{\bf{X}}{{\bf{X}}^ \top }} \right)^{ - 1}}{\bf{X}}{{\bf{T}}^ \top },
\end{equation}
where $\mathbf{T}=\left[ \mathbf{t}\left( 1 \right),\mathbf{t}\left( 2 \right),...,\mathbf{t}\left( {{n}_{max}} \right) \right]$ is the desired output signal. In the training process, the initial reservoir state $\mathbf{x}\left( 0 \right)$ is initialized as a zero matrix and a few warm-up samples are washed out to minimize the impact of the initial zero states.

\begin{remark}
The scaling factor $\lambda $ is important for adjusting the input data to an appropriate range and converting them to a reasonable interval for the activation function. Original ESNs lack the theoretical guidance for setting the scope of $\lambda $, leading to ill-posedness. To address this issue, researchers have suggested various optimization strategies \cite{ref16,ref17,ref18}. However, the optimization process increases the algorithm complexity, creating a trade-off between achieving higher prediction accuracy and maintaining computational efficiency. Therefore, it is crucial to choose a data-dependent and adjustable $\lambda $.    
\end{remark}

\subsection{Stochastic configuration networks}
As a class of randomized learner models, SCNs are innovatively featured by introducing a supervisory mechanism to assign the random parameters in a set of intervals, ensuring the universal approximation property. SCN construction can be easily implemented, and the resulting model has sound performance for both learning and generalization. For more details about SCNs, readers can refer to \cite{ref23}.

Given the objective function $f$, assume that a single-layer feedforward network with $N-1$ hidden layer nodes has been constructed,
\begin{equation}
\label{eq8}
{f_{N - 1}} = \sum\limits_{j = 1}^{N - 1} {{{\bf{\beta }}_j}{g_j}\left( {{\bf{w}}_j^ \top {\bf{u}} + {{\bf{b}}_j}} \right)} \left( {{f_0} = 0,N = 1,2,3...} \right),
\end{equation}
where ${{\mathbf{\beta }}_{j}}={{\left[ {{\beta }_{j,1}},{{\beta }_{j,2}},...{{\beta }_{j,L}} \right]}^{\top }}$ is the output weight of the $j$th hidden layer node, $L$ is the number of output layer nodes, ${{g}_{j}}$ is the activation function, ${{\mathbf{w}}_{j}}$ and ${{\mathbf{b}}_{j}}$ are the input weight and bias, respectively. The residual error between the current model output ${{f}_{N-1}}$ and the expected output ${{f}_{\text{exp}}}$ is expressed as
\begin{equation}
\label{eq9}
{e_{N - 1}} = {f_{{\rm{exp}}}} - {f_{N - 1}} = \left[ {{e_{N - 1,1}},{e_{N - 1,2}},...{e_{N - 1,L}}} \right].
\end{equation}
If ${{\left\| {{e}_{N-1}} \right\|}_{F}}>\varepsilon $, it is necessary to generate a new random basis function ${{g}_{N}}$ under the supervisory mechanism, where $\varepsilon $ is the error tolerance and ${{\left\| \bullet  \right\|}_{F}}$ represents the $F$ norm. Notably, to prevent over-fitting, an additional condition is introduced for stopping the nodes adding and a step size ${{N}_{\text{step}}}$ (${{N}_{\text{step}}}<N$) is used in the following early stopping criterion:
\begin{equation}
\label{eq901}
{\left\| {{e_{{\rm{val}},N - {N_{{\rm{step}}}}}}} \right\|_F} \le {\left\| {{e_{{\rm{val}},N - {N_{{\rm{step}}}} + 1}}} \right\|_F} \le  \ldots  \le {\left\| {{e_{{\rm{val}},N}}} \right\|_F},
\end{equation}
where ${{e}_{\text{val},N}}$ is the validation residual error with $N$ hidden nodes. If Eq. (\ref{eq901}) is satisfied, the number of hidden nodes will be set to $N-{{N}_{\text{step}}}$. The stochastic configuration learning algorithm can be outlined below.

Step 1: Set the parameter scalars \small $\mathbf{\gamma }\text{=}\left\{ {{\lambda }_{\min }}:\Delta \lambda :{{\lambda }_{\max }} \right\}$, \normalsize error tolerance $\varepsilon$, a maximum number of stochastic configurations ${{G}_{\max }}$, a maximum hidden layer size ${{N}_{\max }}$, a step size ${{N}_{\text{step}}}$, and a residual error ${{e}_{0}}={{e}_{N-1}}$.

Step 2: Randomly assign ${{\mathbf{w}}_{N}}$ and ${{\mathbf{b}}_{N}}$ for ${{G}_{\max }}$ times from the adjustable uniform distribution $\left[ -{{\lambda }_{i}},{{\lambda }_{i}} \right]$ to obtain the candidates of basis function ${{g}_{N}}\left( \mathbf{w}_{N}^{\top }\mathbf{u}+{{\mathbf{b}}_{N}} \right)$, which needs to satisfy the following inequality constraint:
\begin{equation}
\label{eq10}
{\left\langle {{e_{N - 1,q}},{g_N}} \right\rangle ^2} \ge b_g^2(1 - r - {\mu _N})\left\| {{e_{N - 1,q}}} \right\|_{}^2,
\end{equation}
where $q = 1,2,...L$, ${{\lambda }_{i}}\in \mathbf{\gamma }$, $0<r<1$, and $\left\{ {{\mu }_{N}} \right\}$ is a nonnegative real sequence that satisfies $\underset{N\to \infty }{\mathop{\lim }}\,{{\mu }_{N}}=0$ and ${{\mu }_{N}}\le (1-r)$, $0<\left\| g \right\|<{{b}_{g}}$.

Step 3: For $q = 1,2,...L$, denoted by
\begin{equation}
\label{eq11}
{\xi _{N,q}} = \frac{{{{\left( {e_{N - 1,q}^ \top {g_N}} \right)}^2}}}{{g_N^ \top {g_N}}} - \left( {1 - {\mu _N} - r} \right)e_{N - 1,q}^ \top e_{N - 1,q}^{}.
\end{equation}
A larger positive value $\xi _{N}^{{}}\text{=}\sum\limits_{q=1}^{L}{{{\xi }_{N,q}}}$ means that the new node is better configured.

Step 4: Update the output weights based on the least square method, that is, 
\begin{equation}
\label{eq12}
\left[ {\beta _1^ * ,\beta _2^ * ,...,\beta _N^ * } \right] = \mathop {\arg \min ||}\limits_{\bf{\beta }} {f_{{\rm{exp}}}} - \sum\limits_{j = 1}^N {{\beta _j}{g_j}} |{|^2}.
\end{equation}

Step 5: Calculate the current residual error ${{e}_{N}}$ and ${{e}_{\text{val},N}}$ for the training and validation sets. If ${{\left\| {{e}_{0}} \right\|}_{F}}\le \varepsilon $, $N\ge {{N}_{\max }}$ or Eq. (\ref{eq901}) is met, the configuration process is completed. Otherwise, renew ${{e}_{0}}:={{e}_{N}}$, update $N:=N+1$, and repeat Step 2-4.

At last, we have $\underset{N\to \infty }{\mathop{\lim }}\,\left\| {{f}_{\text{exp}}}-{{f}_{N}} \right\|=0.$

\section{Recurrent stochastic configuration networks}
This section details the proposed RSCN framework with the algorithm description and theoretical analysis on the echo state property and universal approximation property for offline learning. The architecture of the RSCN is illustrated in Fig.~\ref{fig1} and the construction process is summarized in Algorithm 1. 
\subsection{Algorithm description}
\begin{figure}[htbp]
\centering
\includegraphics[width=3.5in]{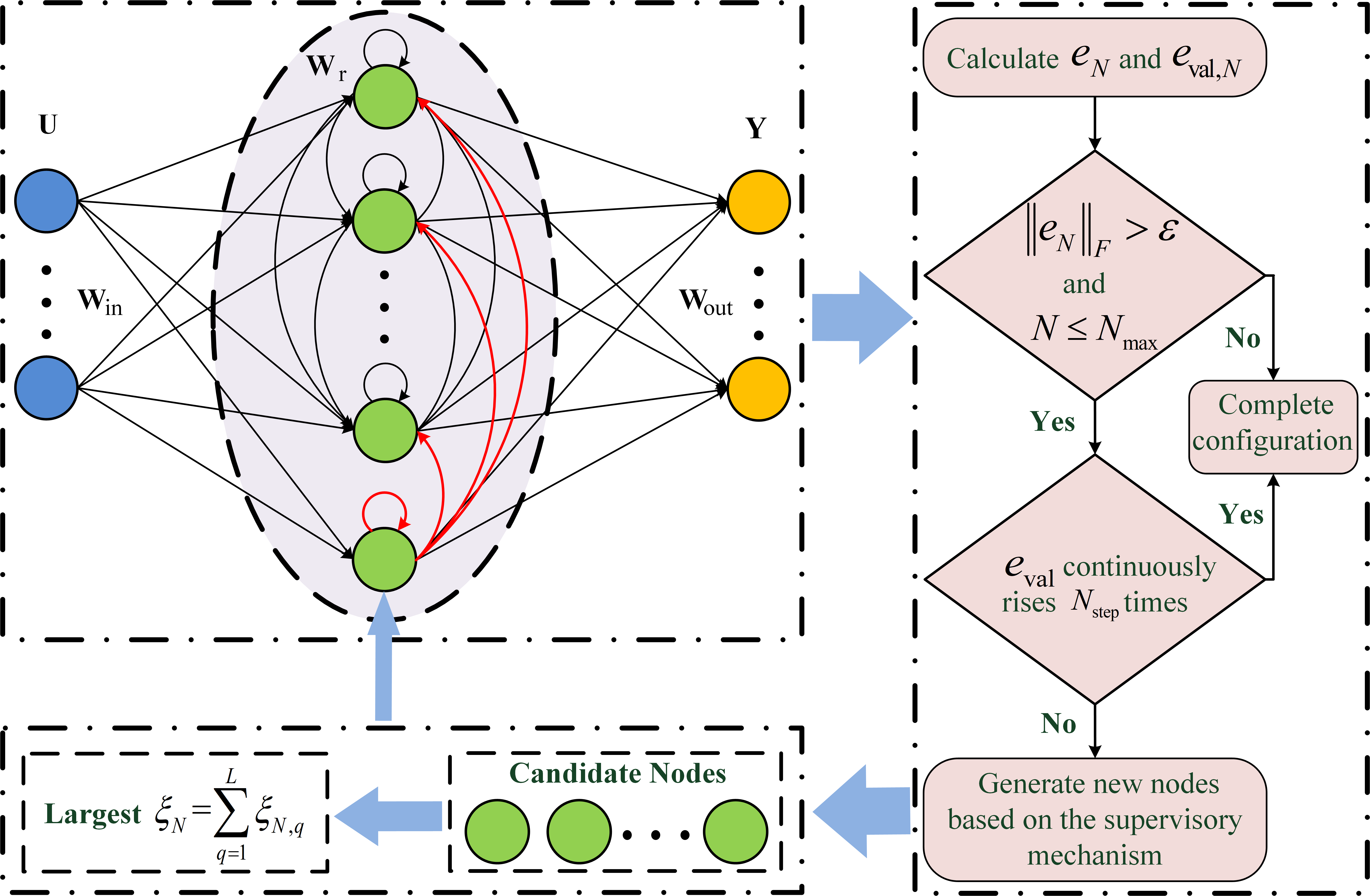}
\caption{Architecture of the recurrent stochastic configuration network.}
\label{fig1}
\end{figure}

Given an RSCN model,
\begin{equation}
\label{eq13}
{\bf{x}}(n) = g({{\bf{W}}_{{\rm{in}}}}{\bf{u}}(n) + {{\bf{W}}_{\mathop{\rm r}\nolimits} }{\bf{x}}(n - 1){\rm{ + }}{\bf{b}}),
\end{equation}
\begin{equation}
\label{eq14}
{\bf{y}}(n) = {{\bf{W}}_{\rm{out}}}\left( {{\bf{x}}(n),{\bf{u}}(n)} \right),
\end{equation}
randomly assign the input weight matrix and biases for the first reservoir node, where ${{\mathbf{W}}_{\operatorname{in},1}}=\left[ \begin{matrix}
   w_{\operatorname{i}\operatorname{n}}^{1,1} & w_{\operatorname{i}\operatorname{n}}^{1,2} & \cdots  & w_{\operatorname{i}\operatorname{n}}^{1,K}  \\
\end{matrix} \right],$ ${{\mathbf{W}}_{\text{r},1}}\text{=}w_{\operatorname{r}}^{1,1}$, ${{\mathbf{b}}_{1}}={{b}_{1}},$ $w_{\operatorname{in}}^{1,j},w_{\operatorname{r}}^{1,1},{{b}_{1}}\in \left[ -\lambda ,\lambda  \right]$. Then, we add nodes in the light of a supervisory mechanism. As shown in Fig.~\ref{fig1}, a special reservoir structure is defined, where only the weights of the new node to the original nodes and itself are assigned, and the weights of other nodes to the new node are set to zero. The feedback matrix can be rewritten as
\begin{small}
\begin{equation}
\label{eq15}
\begin{array}{l}
{{\bf{W}}_{\rm{r},2}}{\rm{ = }}\left[ {\begin{array}{*{20}{c}}
{w_{\mathop{\rm r}\nolimits} ^{1,1}}&0\\
{w_{\mathop{\rm r}\nolimits} ^{2,1}}&{w_{\mathop{\rm r}\nolimits} ^{2,2}}
\end{array}} \right],\\
{{\bf{W}}_{\rm{r},3}}{\rm{ = }}\left[ {\begin{array}{*{20}{c}}
{w_{\mathop{\rm r}\nolimits} ^{1,1}}&0&0\\
{w_{\mathop{\rm r}\nolimits} ^{2,1}}&{w_{\mathop{\rm r}\nolimits} ^{2,2}}&0\\
{w_{\mathop{\rm r}\nolimits} ^{3,1}}&{w_{\mathop{\rm r}\nolimits} ^{3,2}}&{w_{\mathop{\rm r}\nolimits} ^{3,3}}
\end{array}} \right],\\
 \ldots \\
{{\bf{W}}_{{\mathop{\rm r}\nolimits} {\rm{,}}N+1}}{\rm{ = }}\left[ {\begin{array}{*{20}{c}}
{w_{\mathop{\rm r}\nolimits} ^{1,1}}&0& \cdots &0&0\\
{w_{\mathop{\rm r}\nolimits} ^{2,1}}&{w_{\mathop{\rm r}\nolimits} ^{2,2}}& \cdots &0&0\\
 \vdots & \vdots & \vdots & \vdots & \vdots \\
{w_{\mathop{\rm r}\nolimits} ^{N,1}}&{w_{\mathop{\rm r}\nolimits} ^{N,2}}& \cdots &{w_{\mathop{\rm r}\nolimits} ^{N,N}}&0\\
{w_{\mathop{\rm r}\nolimits} ^{N{\rm{ + 1,1}}}}&{w_{\mathop{\rm r}\nolimits} ^{N{\rm{ + 1,2}}}}& \cdots &{w_{\mathop{\rm r}\nolimits} ^{N{\rm{ + }}1,N}}&{w_{\mathop{\rm r}\nolimits} ^{N{\rm{ + }}1,N + 1}}
\end{array}} \right].
\end{array}
\end{equation}
\end{small}
The input weight matrix and biases are denoted as
\begin{equation}
\label{eq16}
\begin{array}{l}
{{\bf{W}}_{{\rm{in}},2}}{\rm{ = }}\left[ {\begin{array}{*{20}{l}}
{w_{{\mathop{\rm i}\nolimits} {\mathop{\rm n}\nolimits} }^{1,1}}&{w_{{\mathop{\rm i}\nolimits} {\mathop{\rm n}\nolimits} }^{1,2}}& \cdots &{w_{{\mathop{\rm i}\nolimits} {\mathop{\rm n}\nolimits} }^{1,K}}\\
{w_{{\mathop{\rm i}\nolimits} {\mathop{\rm n}\nolimits} }^{2,1}}&{w_{{\mathop{\rm i}\nolimits} {\mathop{\rm n}\nolimits} }^{2,2}}& \cdots &{w_{{\mathop{\rm i}\nolimits} {\mathop{\rm n}\nolimits} }^{2,K}}
\end{array}} \right],\\
 \ldots \\
{{\bf{W}}_{{\rm{in}},N + 1}}{\rm{ = }}\left[ {\begin{array}{*{20}{c}}
{w_{{\mathop{\rm i}\nolimits} {\mathop{\rm n}\nolimits} }^{1,1}}&{w_{{\mathop{\rm i}\nolimits} {\mathop{\rm n}\nolimits} }^{1,2}}& \cdots &{w_{{\mathop{\rm i}\nolimits} {\mathop{\rm n}\nolimits} }^{1,K}}\\
{w_{{\mathop{\rm i}\nolimits} {\mathop{\rm n}\nolimits} }^{2,1}}&{w_{{\mathop{\rm i}\nolimits} {\mathop{\rm n}\nolimits} }^{2,2}}& \cdots &{w_{{\mathop{\rm i}\nolimits} {\mathop{\rm n}\nolimits} }^{2,K}}\\
 \vdots & \vdots & \vdots & \vdots \\
{w_{{\mathop{\rm i}\nolimits} {\mathop{\rm n}\nolimits} }^{N,1}}&{w_{{\mathop{\rm i}\nolimits} {\mathop{\rm n}\nolimits} }^{N,2}}& \cdots &{w_{{\mathop{\rm i}\nolimits} {\mathop{\rm n}\nolimits} }^{N,K}}\\
{w_{{\mathop{\rm i}\nolimits} {\mathop{\rm n}\nolimits} }^{N + 1,1}}&{w_{{\mathop{\rm i}\nolimits} {\mathop{\rm n}\nolimits} }^{N + 1,2}}& \cdots &{w_{{\mathop{\rm i}\nolimits} {\mathop{\rm n}\nolimits} }^{N + 1,K}}
\end{array}} \right],\\
{{\bf{b}}_2} = {\left[ {{b_1},{b_2}} \right]^ \top }, \ldots ,{{\bf{b}}_{N + 1}} = {\left[ {{b_1}, \ldots {b_{N + 1}}} \right]^ \top }.
\end{array}
\end{equation}

Suppose we have built an RSCN with $N$ reservoir nodes, when adding a new node, set the initial state of the new node to zero, ${{\mathbf{x}}_{N+1}}(n-1)=$${{\left[ {{x}_{1}}(n-1),\ldots ,{{x}_{N}}(n-1),0 \right]}^{\top }}$. The reservoir state is given by
\begin{equation}
\label{eq17}
\begin{array}{l}
{{\bf{x}}_{N{\rm{ + }}1}}(n) = g\left( {{{\bf{W}}_{{\mathop{\rm in}\nolimits} ,N + 1}}{\bf{u}}(n)} \right.\\
{\kern 1pt} {\kern 1pt} {\kern 1pt} {\kern 1pt} {\kern 1pt} {\kern 1pt} {\kern 1pt} {\kern 1pt} {\kern 1pt} {\kern 1pt} {\kern 1pt} {\kern 1pt} {\kern 1pt} {\kern 1pt} {\kern 1pt} {\kern 1pt} {\kern 1pt} {\kern 1pt} {\kern 1pt} {\kern 1pt} {\kern 1pt} {\kern 1pt} {\kern 1pt} {\kern 1pt} {\kern 1pt} {\kern 1pt} {\kern 1pt} {\kern 1pt} {\kern 1pt} {\kern 1pt} {\kern 1pt} {\kern 1pt} {\kern 1pt} {\kern 1pt} {\kern 1pt} {\kern 1pt} {\kern 1pt} {\kern 1pt} {\kern 1pt} {\kern 1pt} {\kern 1pt} {\kern 1pt} {\kern 1pt} {\kern 1pt} {\kern 1pt} {\kern 1pt} {\kern 1pt} {\kern 1pt} {\kern 1pt} {\kern 1pt} {\kern 1pt} {\kern 1pt} {\kern 1pt} {\kern 1pt} {\kern 1pt} {\kern 1pt}  + \left. {{{\bf{W}}_{{\mathop{\rm r}\nolimits} ,N + 1}}{{\bf{x}}_{N + 1}}(n - 1) + {{\bf{b}}_{N + 1}}} \right)\\
{\kern 1pt} {\kern 1pt} {\kern 1pt} {\kern 1pt} {\kern 1pt} {\kern 1pt} {\kern 1pt} {\kern 1pt} {\kern 1pt} {\kern 1pt} {\kern 1pt} {\kern 1pt} {\kern 1pt} {\kern 1pt} {\kern 1pt} {\kern 1pt} {\kern 1pt} {\kern 1pt} {\kern 1pt} {\kern 1pt} {\kern 1pt} {\kern 1pt} {\kern 1pt} {\kern 1pt} {\kern 1pt} {\kern 1pt} {\kern 1pt} {\kern 1pt} {\kern 1pt} {\kern 1pt} {\kern 1pt} {\kern 1pt} {\kern 1pt} {\kern 1pt} {\kern 1pt} {\kern 1pt} {\kern 1pt}  = g({\bf{x}}_{{\mathop{\rm in}\nolimits} ,N + 1}^{}(n) + {\bf{x}}_{{\rm{r}},N + 1}^{}(n - 1)),
\end{array}
\end{equation}
where 
\begin{scriptsize}
\begin{equation}
\label{eq18}
\begin{array}{l}
{\bf{W}}_{{\mathop{\rm i}\nolimits} {\mathop{\rm n}\nolimits} ,N + 1}^{}{\bf{u}}(n){\rm{ + }}{{\bf{b}}_{N + 1}} = \\
\left[ {\begin{array}{*{20}{c}}
{w_{{\mathop{\rm i}\nolimits} {\mathop{\rm n}\nolimits} }^{1,1}}&{w_{{\mathop{\rm i}\nolimits} {\mathop{\rm n}\nolimits} }^{1,2}}& \cdots &{w_{{\mathop{\rm i}\nolimits} {\mathop{\rm n}\nolimits} }^{1,K}}\\
{w_{{\mathop{\rm i}\nolimits} {\mathop{\rm n}\nolimits} }^{2,1}}&{w_{{\mathop{\rm i}\nolimits} {\mathop{\rm n}\nolimits} }^{2,2}}& \cdots &{w_{{\mathop{\rm i}\nolimits} {\mathop{\rm n}\nolimits} }^{2,K}}\\
 \vdots & \vdots & \vdots & \vdots \\
{w_{{\mathop{\rm i}\nolimits} {\mathop{\rm n}\nolimits} }^{N,1}}&{w_{{\mathop{\rm i}\nolimits} {\mathop{\rm n}\nolimits} }^{N,2}}& \cdots &{w_{{\mathop{\rm i}\nolimits} {\mathop{\rm n}\nolimits} }^{N,K}}\\
{w_{{\mathop{\rm i}\nolimits} {\mathop{\rm n}\nolimits} }^{N + 1,1}}&{w_{{\mathop{\rm i}\nolimits} {\mathop{\rm n}\nolimits} }^{N + 1,2}}& \cdots &{w_{{\mathop{\rm i}\nolimits} {\mathop{\rm n}\nolimits} }^{N + 1,K}}
\end{array}} \right]{\bf{u}}(n) + \left[ {\begin{array}{*{20}{c}}
{{b_1}}\\
{{b_2}}\\
 \vdots \\
{{b_N}}\\
{{b_{N + 1}}}
\end{array}} \right]\\
{\kern 1pt} {\rm{ = }}{\kern 1pt} {\kern 1pt} {\bf{x}}_{{\mathop{\rm in}\nolimits} ,N + 1}^{}(n - 1) \in {{\mathbb{R}}^{N+1}},
\end{array}
\end{equation}
\end{scriptsize}
\begin{scriptsize}
\begin{equation}
\label{eq19}
\begin{array}{l}
{{\bf{W}}_{{\mathop{\rm r}\nolimits} ,N + 1}}{{\bf{x}}_{N + 1}}(n - 1){\rm{ = }}\\
\left[ {\begin{array}{*{20}{c}}
{w_{\mathop{\rm r}\nolimits} ^{1,1}}&0& \cdots &0&0\\
{w_{\mathop{\rm r}\nolimits} ^{2,1}}&{w_{\mathop{\rm r}\nolimits} ^{2,2}}& \cdots &0&0\\
 \vdots & \vdots & \vdots & \vdots & \vdots \\
{w_{\mathop{\rm r}\nolimits} ^{N,1}}&{w_{\mathop{\rm r}\nolimits} ^{N,2}}& \cdots &{w_{\mathop{\rm r}\nolimits} ^{N,N}}&0\\
{w_{\mathop{\rm r}\nolimits} ^{N{\rm{ + 1,1}}}}&{w_{\mathop{\rm r}\nolimits} ^{N{\rm{ + 1,2}}}}& \cdots &{w_{\mathop{\rm r}\nolimits} ^{N{\rm{ + }}1,N}}&{w_{\mathop{\rm r}\nolimits} ^{N{\rm{ + }}1,N + 1}}
\end{array}} \right]\left[ {\begin{array}{*{20}{c}}
{{x_1}(n - 1)}\\
{{x_2}(n - 1)}\\
 \vdots \\
{{x_N}(n - 1)}\\
0
\end{array}} \right]\\
{\kern 1pt} {\rm{ = }}{\left[ {x_1^{}\left( {n - 1} \right),...,x_N^{}\left( {n - 1} \right),x_{N + 1}^{}\left( {n - 1} \right)} \right]^ \top }{\rm{ = }}{\kern 1pt} {\kern 1pt} {\kern 1pt} {\kern 1pt} {\bf{x}}_{{\rm{r}},N + 1}^{}(n - 1) \in {{\mathbb{R}}^{N+1}}.
\end{array}
\end{equation}
\end{scriptsize}
Given the input signals $\mathbf{U}\text{=}\left[ \mathbf{u}(1),\mathbf{u}(2),\ldots ,\mathbf{u}({{n}_{\max }}) \right]$, the error between the model output and the desired output is defined as $e_N=\mathbf{Y}-\mathbf{T}$. When ${{\left\| {{e}_{N}} \right\|}_{F}}$ does not meet the preset threshold and Eq. (\ref{eq901}) is not satisfied, it is necessary to add nodes under the supervisory mechanism. Seek the random basis function ${g_{N{\rm{ + }}1}}$ satisfying the following inequality:
\begin{equation}
\label{eq20}
{\left\langle {e_{N,q}^{},{g_{N{\rm{ + }}1}}} \right\rangle ^2} \ge b_g^2(1 - r_i - {\mu _{N + 1}})\left\| {{e_{N,q}}} \right\|_{}^2,
\end{equation}
where $q = 1,2,...L$, and $\left\{ {{\mu }_{N+1}} \right\}$ is a nonnegative real sequence that satisfies $\underset{N\to \infty }{\mathop{\lim }}\,{{\mu }_{N+1}}=0$, ${{\mu }_{N+1}}\le (1-r_i)$, ${r_i} \in \left\{ {{r_1}, \ldots ,{r_t}} \right\}$, $(0 < {r_i} < 1)$, and $0<\left\| g \right\|<{{b}_{g}}$. In addition, to facilitate the implementation of RSCNs, ${\xi _{N+1,q}}$ is calculated for each candidate node according to Eq. (\ref{eq11}). If ${\xi _{N+1,q}}>0$, select the candidate node with the largest ${{\xi }_{N+1}}=\sum\limits_{q=1}^{L}{{{\xi }_{N+1,q}}}$ as the optimal adding node. Else if $i=t-1$ then set a new $\lambda $, $i=1$ and repeat. Else repeat with $i: = i + 1$.

The output weight is determined by the least square method, that is,
\begin{equation}
\label{eq21}
\begin{array}{*{20}{l}}
{{\bf{W}}_{{\rm{out}},N{\rm{ + }}1}^{}{\rm{ = }}\left[ {{\bf{w}}_{{\rm{out}},1}^{},{\bf{w}}_{{\rm{out}},2}^{},...,{\bf{w}}_{{\rm{out}},N{\rm{ + }}1{\rm{ + }}K}^{}} \right]}\\
{{\kern 1pt} {\kern 1pt} {\kern 1pt} {\kern 1pt}{\kern 1pt} {\kern 1pt} {\kern 1pt} {\kern 1pt} {\kern 1pt} {\kern 1pt} {\kern 1pt} {\kern 1pt} {\kern 1pt} {\kern 1pt} {\kern 1pt} {\kern 1pt} {\kern 1pt} {\kern 1pt} {\kern 1pt} {\kern 1pt} {\kern 1pt} {\kern 1pt} {\kern 1pt} {\kern 1pt} {\kern 1pt} {\kern 1pt} {\kern 1pt} {\kern 1pt} {\kern 1pt} {\kern 1pt} {\kern 1pt} {\kern 1pt} {\kern 1pt} {\kern 1pt} {\kern 1pt} {\kern 1pt} {\kern 1pt} {\kern 1pt} {\kern 1pt} {\kern 1pt} {\kern 1pt}  = \mathop {\arg \min }\limits_{{\bf{W}}_{{\rm{out}}}^{}} \left\| {{\bf{T}} - {\bf{W}}_{{\rm{out}}}^{}{{\bf{X}}_{N + 1}}} \right\|_{}^2,}
\end{array}
\end{equation}
where \small ${{\mathbf{X}}_{N+1}}\text{=}\left[ \left( {{\mathbf{x}}_{N+1}}\left( 1 \right),\mathbf{u}\left( 1 \right) \right),\ldots ,\left( {{\mathbf{x}}_{N+1}}\left( {{n}_{\max }} \right),\mathbf{u}\left( {{n}_{\max }} \right) \right) \right]$. \normalsize Calculate the residual error ${{e}_{N+1}}$ and update $N=N+1$. Continue to add nodes until ${{\left\| e \right\|}_{F}}<\varepsilon$ or $N\ge {{N}_{\max }}$ or Eq. (\ref{eq901}) is met, where ${{N}_{\max }}$ is the maximum reservoir size. 
\begin{remark}
The model output can be regarded as a linear combination of reservoir output and input direct link. We can obtain the globally optimal $\mathbf{W}_{\text{out}}^{{}}$ using the least squares method. Further research is necessary for the system that requires accurate calculation of the individual weights corresponding to two components.
\end{remark}
\begin{algorithm}[t]\footnotesize 
\caption{Recurrent stochastic configuration}
\KwIn{Training inputs $\mathbf{U}\text{=}\left[ \mathbf{u}(1),\ldots ,\mathbf{u}({{n}_{\max }}) \right]$, training outputs ${\bf{T}} = {\left[ {{\bf{t}}\left( 1 \right),\ldots ,{\bf{t}}\left( {{n_{max}}} \right)} \right]}$, validation inputs ${{\bf{U}}_{{\rm{val}}}}{\rm{ = }}\left[ {{{\bf{u}}_{{\rm{val}}}}(1),{{\bf{u}}_{{\rm{val}}}}(2), \ldots } \right]$, validation outputs ${{\bf{T}}_{{\rm{val}}}}{\rm{ = }}\left[ {{{\bf{t}}_{{\rm{val}}}}(1),{{\bf{t}}_{{\rm{val}}}}(2), \ldots } \right]$, a initial reservoir size $N$, a maximum reservoir size ${N_{\max }}$, a step size ${{N}_{\text{step}}}$, a training error threshold $\varepsilon$, the positive scalars $\mathbf{\gamma }\text{=}\left\{ {{\lambda }_{1}},{{\lambda }_{2}},...,{{\lambda }_{\max }} \right\}$, the contractive factors $\left\{ {{r_1}, \ldots ,{r_t}} \right\},(0 < {r_i} < 1)$, and a maximum number of stochastic configurations ${{G}_{\max }}$.} 
\KwOut{RSCN}
Initialization: Randomly assign ${{\mathbf{W}}_{\operatorname{in},N}}$, ${{\bf{b}}_N}$, and ${{\bf{W}}_{{\rm{r}},{{N}}}}$ according to the sparsity of the reservoir from $\left[ -\lambda ,\lambda  \right]$, and calculate the model output $\mathbf{Y}$, current residual error ${e}_{N}$ and ${{e}_{\text{val},N}}$, let ${{e}_{0}}={{e}_{N}}$, $\mathbf{\Omega },\mathbf{D}:=\left[ {\kern 1pt} {\kern 1pt} {\kern 1pt} \right]$.\\
\While{$N\le {{N}_{\text{max}}}$ AND ${{\left\| e_{0}^{{}} \right\|}_{F}}>\varepsilon $,}{
   \If{Eq. (\ref{eq901}) is not met}{
    \For{$\lambda \in \mathbf{\gamma }$}{
        \For{$k=1,2,...,{{G}_{\max }}$}{
            Randomly assign ${b_{N + 1}}$, ${\bf{w}}_{{\mathop{\rm i}\nolimits} {\mathop{\rm n}\nolimits} }^{N + 1} = \left[ \begin{matrix} w_{\operatorname{i}\operatorname{n}}^{N + 1,1} & \cdots  & w_{\operatorname{i}\operatorname{n}}^{N + 1,K}\end{matrix} \right],$ and
          ${\bf{w}}_{{\mathop{\rm r}\nolimits} }^{N + 1} =\left[ \begin{matrix} w_{\operatorname{r}}^{N + 1,1} & \cdots  & w_{\operatorname{r}}^{N + 1,N + 1} \end{matrix} \right]$ from $\left[ -\lambda ,\lambda  \right]$;\\
            Construct $\mathbf{W}_{\operatorname{i}\operatorname{n},N+1}^{{}}$ and $\mathbf{W}_{\text{r},N+1}^{{}}$;\\
            Calculate $\mathbf{x}_{\operatorname{in},N+1}^{{}}(n-1)$ and $\mathbf{x}_{\text{r},N+1}^{{}}(n-1)$ based on Eq. (\ref{eq18}) and Eq. (\ref{eq19});\\
            Calculate ${{g}_{N+1}}$ and ${{\xi }_{N\text{+}1,q}}$;\\
            Set ${{\mu }_{N\text{+}1}}\text{=}(1-r_i)/\left( N\text{+}1 \right)$;\\
            \If{$\min \left\{ {{\xi }_{N\text{+}1,1}},{{\xi }_{N\text{+}1,2}},...,{{\xi }_{N\text{+1},L}} \right\}\ge 0$}{
                Save $\mathbf{w}_{\operatorname{i}\operatorname{n}}^{N+1},$ ${{b}_{N+1}}$ and $\mathbf{w}_{\text{r}}^{N+1}$ in $\mathbf{D}$, $\xi _{N\text{+}1}^{{}}\text{=}\sum\limits_{q=1}^{L}{{{\xi }_{N\text{+}1,q}}}$ in $\mathbf{\Omega }$;\\
            \Else{Go back to Step 5}
            } 
        }  
        \If{$\mathbf{D}$ is not empty}{
            Find $\mathbf{w}_{\operatorname{i}\operatorname{n}}^{\left( N+1 \right)*}$, $b_{N+1}^{*}$, and $\mathbf{w}_{\text{r}}^{\left( N+1 \right)*}$ that maximize $\xi _{N\text{+}1}^{{}}$ in $\mathbf{\Omega }$, construct $\mathbf{W}_{\text{in},N+1}^{*}$, $\mathbf{W}_{\text{r},N+1}^{*}$, and $\mathbf{b}_{N+1}^{*}$, calculate $g_{N+1}^{*}$, and get ${{\mathbf{X}}_{N\text{+}1}}$;\\
            \textbf{Break} (go to Step 30)\\
            \Else{
                \While{$i<t$}{Set $i=i+1$ and take $r_i$;
                } 
                Go back to Step 5;
            }
        }
        Go back to Step 4;
    } 
    Calculate $\mathbf{W}_{\operatorname{out},N\text{+}1}^{*}$ according to Eq. (\ref{eq21});\\
    Calculate $e_{N\text{+}1}^{{}}\text{=}e_{N}^{{}}-\mathbf{w}_{\operatorname{out},N+1}^{*}g_{N+1}^{*}$ and ${{e}_{\text{val},N+1}}$;\\
    Update ${{e}_{0}}:=e_{N+1}^{{}}$, $N:=N+1$;\\
  \Else{Set the optimal reservoir size to $N:=N-{{N}_{\text{step}}}$ and obtain the current network parameters;\\
  \textbf{Break} (go to Step 39)\\}  
  }
 }  
 \textbf{Return:} $\mathbf{W}_{\text{in},N}^{*}$, $\mathbf{b}_{N}^{*}$, $\mathbf{W}_{\text{r},N}^{*}$, and $\mathbf{W}_{\text{out},N}^{*}$.
\end{algorithm}
\subsection{The echo state property of RSCN}
The echo state property is an intrinsic characteristic that distinguishes ESN from other networks \cite{ref21}. As the length of the input sequence approaches infinity, the error between two reservoir states ${\bf{x}}^1\left( n \right)$ and ${\bf{x}}^2\left( n \right)$, driven by the same input sequence but starting from different initial conditions, tends to zero. And the dependence of ${\bf{x}}\left( n \right)$ on the initial state ${\bf{x}}\left( 0 \right)$ diminishes over time, indicating that the current state only relies on historical inputs and is independent of the initial state.

Let ${{\sigma }_{\max }}\left( {{\mathbf{W}}_{\text{r}}} \right)$ denote the maximal singular value of ${{\mathbf{W}}_{\text{r}}}$. The spectral radius of ${{\mathbf{W}}_{\text{r}}}$ is set to $\rho \left( {{\mathbf{W}}_{\text{r}}} \right)=\max \left( \left| {{\theta }_{1}} \right|,\left| {{\theta }_{2}} \right|,\ldots  \right)$, where ${{\theta }_{i}}$ is the eigenvalue of ${{\mathbf{W}}_{\text{r}}}$. Observe that the largest singular value of ${{\mathbf{W}}_{\text{r}}}$ can be scaled with $\alpha$, that is, ${{\sigma }_{\max }}\left( \alpha {{\mathbf{W}}_{\text{r}}} \right)=\alpha {{\sigma }_{\max }}\left( {{\mathbf{W}}_{\text{r}}} \right)$. From \cite{ref351}, we know that for each square matrix ${{\mathbf{W}}_{\text{r}}}$, $\rho \left( {{\mathbf{W}}_{\text{r}}} \right)\le {{\sigma }_{\max }}\left( {{\mathbf{W}}_{\text{r}}} \right)$. To ensure the echo state property, the feedback matrix ${\bf{W}}_{{\rm{r}},{{N + 1}}}$ is scaled, that is,
\begin{equation}
\label{eq22}
{{\bf{W}}_{{\rm{r}},{{N + 1}}}} \leftarrow \frac{\alpha }{{\rho \left( {{{\bf{W}}_{{\rm{r}},{{N + 1}}}}} \right)}}{{\bf{W}}_{{\rm{r}},{{N + 1}}}},
\end{equation}
where $0<\alpha<1$ represents the scaling factor. Substitute ${{\bf{W}}_{{\rm{r}},{{N + 1}}}}$ obtained from Eq. (\ref{eq15}) into Eq. (\ref{eq22}) and seek for the nodes that satisfy the inequality constraints in Eq. (\ref{eq20}). \\

\textbf{Theorem 1.} The echo state property of RSCN can be ensured if $\alpha$ is chosen as
\begin{equation}
\label{eq23}
0 < \alpha  < \frac{{\rho \left( {{{\bf{W}}_{{\rm{r}},{{N + 1}}}}} \right)}}{{{\sigma _{\max }}\left( {{{\bf{W}}_{{\rm{r}},{{N + 1}}}}} \right)}}.
\end{equation}

\textbf{Proof.} Randomly assign two initial states ${\bf{x}}_{N + 1}^1\left( 0 \right)$ and ${\bf{x}}_{N + 1}^2\left( 0 \right)$. After the system evolves for a period of time, the difference between the reservoir states is given by
\begin{equation}
\label{eq24}
\begin{array}{*{20}{l}}
{{{\left\| {{\bf{x}}_{N + 1}^1\left( n \right) - {\bf{x}}_{N + 1}^2\left( n \right)} \right\|}^2}}\\
{ = \left\| {g({{\bf{W}}_{{\rm{in}},N + 1}}{\bf{u}}(n) + {{\bf{W}}_{{\rm{r}},N + 1}}{\bf{x}}_{N + 1}^1(n - 1) + {{\bf{b}}_{N + 1}})} \right.}\\
{  - {{\left. {g({{\bf{W}}_{{\rm{in}},N + 1}}{\bf{u}}(n) + {{\bf{W}}_{{\rm{r}},N + 1}}{\bf{x}}_{N + 1}^2(n - 1) + {{\bf{b}}_{N + 1}})} \right\|}^2},}
\end{array}
\end{equation}
where $g$ is the sigmoid or tanh function. According to the Lipschitz condition, we have
\begin{equation}
\label{eq25}
\begin{array}{*{20}{l}}
{{{\left\| {{\bf{x}}_{N + 1}^1\left( n \right) - {\bf{x}}_{N + 1}^2\left( n \right)} \right\|}^2}}\\
{ \le \max \left( {\left| {g'} \right|} \right){{\left\| {{{\bf{W}}_{{\rm{r}},N + 1}}{\bf{x}}_{N + 1}^1(n - 1) - {{\bf{W}}_{{\rm{r}},N + 1}}{\bf{x}}_{N + 1}^2(n - 1)} \right\|}^2}}\\
{ \le \left\| {{{\bf{W}}_{{\rm{r}},N + 1}}} \right\|{{\left\| {{\bf{x}}_{N + 1}^1\left( {n - 1} \right) - {\bf{x}}_{N + 1}^2\left( {n - 1} \right)} \right\|}^2}}\\
{ \le \left\| {{{\bf{W}}_{{\rm{r}},N + 1}}} \right\|_{}^2{{\left\| {{\bf{x}}_{N + 1}^1\left( {n - 2} \right) - {\bf{x}}_{N + 1}^2\left( {n - 2} \right)} \right\|}^2}}\\
{ \le  \ldots }{ \le \left\| {{{\bf{W}}_{{\rm{r}},N + 1}}} \right\|_{}^n{{\left\| {{\bf{x}}_{N + 1}^1\left( 0 \right) - {\bf{x}}_{N + 1}^2\left( 0 \right)} \right\|}^2},}
\end{array}
\end{equation}
where ${g'}$ is the first order derivative of $g$. Note that
\begin{equation}
\label{eq26}
\begin{array}{l}
{\sigma _{\max }}\left( {\frac{\alpha }{{\rho \left( {{{\bf{W}}_{{\rm{r}},N{\rm{ + 1}}}}} \right)}}{{\bf{W}}_{{\rm{r}},N{\rm{ + 1}}}}} \right) = \frac{\alpha }{{\rho \left( {{{\bf{W}}_{{\rm{r}},N{\rm{ + 1}}}}} \right)}}{\sigma _{\max }}\left( {{{\bf{W}}_{{\rm{r}},N{\rm{ + 1}}}}} \right)\\
 < \frac{{\rho \left( {{{\bf{W}}_{{\rm{r}},N{\rm{ + 1}}}}} \right)}}{{{\sigma _{\max }}\left( {{{\bf{W}}_{{\rm{r}},N{\rm{ + 1}}}}} \right)}} \times \frac{1}{{\rho \left( {{{\bf{W}}_{{\rm{r}},N{\rm{ + 1}}}}} \right)}}{\sigma _{\max }}\left( {{{\bf{W}}_{{\rm{r}},N{\rm{ + 1}}}}} \right) = 1.
\end{array}
\end{equation}
Since the Euclidean norm of the matrix is equal to the maximum singular value, we have $\left\| {{\mathbf{W}}_{\text{r},N+1}} \right\|<1$ and $\underset{n\to \infty }{\mathop{\lim }}\,\left\| {{\mathbf{W}}_{\operatorname{r},N+1}} \right\|_{{}}^{n}=0$. Thus, we have $\mathop {\lim }\limits_{n \to \infty } \left\| {{\bf{x}}_{N + 1}^1\left( n \right) - } \right.$ $\left. {{\bf{x}}_{N + 1}^2\left( n \right)} \right\| = 0$, which completes the proof.

\begin{remark}
The echo state property ensures the model's stable response to the input data. During the initial phase, the model output can be influenced by its initial state. To get a reasonable and logical response to the input signal, the network must be run for a sufficient duration, requiring a period of input sequence to warm up the model and stabilize its reservoir state. The length of the warm-up period will vary based on the complexity of the data and the dynamic characteristics of the network. Once the warm-up phase is complete, subsequent input data will primarily affect the reservoir's state, independent of the initial condition. As a result, the model can then consistently capture and predict temporal relationships in the data, demonstrating stable performance across various applications, particularly in time series analysis and dynamic system modelling.
\end{remark}

\subsection{The universal approximation property for offline learning}
RSCNs possess the capability to approximate any complex function, offering a significant advantage in modelling uncertain and intricate industrial systems. This subsection presents a theoretical proof of the universal approximation property of the proposed RSCN for offline learning.\\

\textbf{Theorem 2.} Suppose that span($\Gamma$) is dense in ${{L}_{2}}$ space, for $b_{g}\in {{\mathbb{R}}^{+}}$ , $\forall g\in \Gamma ,0<\left\| g \right\|<{{b}_{g}}$. Given $0<r<1$ and a nonnegative real sequence $\left\{ {{\mu }_{N+1}} \right\}$ satisfies $\underset{N\to \infty }{\mathop{\lim }}\,{{\mu }_{N\text{+}1}}=0$ and ${{\mu }_{N\text{+}1}}\le (1-r)$. For $N=1,2...$, $q=1,2,...,L$, define
\begin{equation}
\label{eq36}
{\delta _{N{\rm{ + 1}}}} = \sum\limits_{q = 1}^L {\delta _{N{\rm{ + }}1,q}^{}} {\kern 1pt} {\kern 1pt} {\kern 1pt} {\kern 1pt} ,{\kern 1pt} {\kern 1pt} {\kern 1pt} \delta _{N{\rm{ + }}1,q}^{} = (1 - r - {\mu _{N{\rm{ + }}1}})\left\| {e_{N,q}^{}} \right\|_{}^2.
\end{equation}
If $g_{N\text{+}1}$ satisfies the following inequality constraints:
\begin{equation}
\label{eq37}
{\left\langle {e_{N,q}^{},{g_{N{\rm{ + }}1}}} \right\rangle ^2} \ge b_g^2\delta _{N{\rm{ + }}1,q}^{},q = 1,2,...L,
\end{equation}
and the output weight is evaluated by the least squares method, that is,
\begin{equation}
\label{eq38}
{\left[ {{\bf{w}}_{{\rm{out}},1}^*,...,{\bf{w}}_{{\rm{out}},N + 1}^*} \right] = \mathop {\arg \min }\limits_{{\bf{W}}_{{\rm{out}},N + 1}^{}} \left\| {{\bf{T}} - {\bf{W}}_{{\rm{out}},N + 1}^{}{{\bf{X}}_{N + 1}}} \right\|_{}^2.}
\end{equation}
Then, we have ${{\lim }_{N\to \infty }}\left\| e_{N\text{+}1} \right\|=0$. 

\textbf{Proof.} By referring to the proof in \cite{ref23}, the conclusion follows directly. Indeed, it is sufficient to estimate the term $\left\| {e_{N{\rm{ + }}1}^{}} \right\|_{}^2 - \left( {r + {\mu _{N{\rm{ + }}1}}} \right)\left\| {e_N^{}} \right\|_{}^2$ for a special output weight:
\begin{equation}
\label{eq381}
{\bf{w}}_{{\mathop{\rm out}\nolimits} ,N + 1,q}^* = \frac{{\left\langle {e_{N,q}^{},g_{N{\rm{ + }}1}^{}} \right\rangle }}{{\left\| {g_{N{\rm{ + }}1}^{}} \right\|_{}^2}},q = 1,2,...,L.
\end{equation}
With simple computation, we have
\begin{equation}
\label{eq39}
\begin{array}{l}
\left\| {e_{N{\rm{ + }}1}^{}} \right\|_{}^2 - \left( {r + {\mu _{N{\rm{ + }}1}}} \right)\left\| {e_N^{}} \right\|_{}^2\\
= \sum\limits_{q = 1}^L {\left\langle {e_{N,q}^{} - {\bf{w}}_{{\mathop{\rm out}\nolimits} ,N + 1}^*g_{N + 1}^{},e_{N,q}^{} - {\bf{w}}_{{\mathop{\rm out}\nolimits} ,N + 1}^*g_{N + 1}^{}} \right\rangle } \\
 - \sum\limits_{q = 1}^L {\left( {r + {\mu _{N{\rm{ + }}1}}} \right)\left\langle {e_{N,q}^{},e_{N,q}^{}} \right\rangle } \\
= \left( {1 - r - {\mu _{N{\rm{ + }}1}}} \right)\left\| {e_N^{}} \right\|_{}^2 - \frac{{\sum\limits_{q = 1}^L {{{\left\langle {e_{N,q}^{},g_{N + 1}^{}} \right\rangle }^2}} }}{{\left\| {g_{N + 1}^{}} \right\|_{}^2}}\\
= {\delta _{N{\rm{ + 1}}}} - \frac{{\sum\limits_{q = 1}^L {{{\left\langle {e_{N,q}^{},g_{N + 1}^{}} \right\rangle }^2}} }}{{\left\| {g_{N + 1}^{}} \right\|_{}^2}} \le {\delta _{N{\rm{ + 1}}}} - \frac{{\sum\limits_{q = 1}^L {{{\left\langle {e_{N,q}^{},g_{N + 1}^{}} \right\rangle }^2}} }}{{b_g^2}} \le 0.
\end{array}
\end{equation}
Therefore, the following inequality holds:
\begin{equation}
\label{eq40}
\left\| {e_{N{\rm{ + }}1}^{}} \right\|_{}^2 \le r\left\| {e_N^{}} \right\|_{}^2 + {\gamma _{N + 1}},
\end{equation}
where ${{\gamma _{N + 1}} = {\mu _{N{\rm{ + }}1}}\left\| {e_N^{}} \right\|_{}^2 }$. Note that ${{\lim }_{N\to \infty }}{{\gamma }_{N+1}}=0$. From Eq. (\ref{eq40}), we have $\underset{N\to \infty }{\mathop{\lim }}\,\left\| {{e}_{N\text{+}1}} \right\|_{{}}^{2}=0$, which completes the proof. 

\begin{remark}
The universal approximation performance of the RSCN involves a condition, $e_{N{\rm{ + }}1}^{}{\rm{ = }}e_N^{} - {\bf{w}}_{{\rm{out}},N + 1}^*g_{N + 1}^{}$. This implies that when adding a new node, the model output of the first $N$ nodes remains unchanged. If we construct the feedback matrix in a general form, that is,
\begin{small}
 \begin{equation}
\label{eq42}
\begin{array}{l}
{\bf{W}}_{{\rm{r}},N + 1}^*{{\bf{x}}_{N + 1}}(n){\rm{ = }}\\
\left[ {\begin{array}{*{20}{c}}
{w_{\rm{r}}^{1,1}}&{w_{\rm{r}}^{1,2}}& \cdots &{w_{\rm{r}}^{1,N}}&{w_{\rm{r}}^{1,N{\rm{ + }}1}}\\
{w_{\rm{r}}^{2,1}}&{w_{\rm{r}}^{2,2}}& \cdots &{w_{\rm{r}}^{2,N}}&{w_{\rm{r}}^{2,N{\rm{ + }}1}}\\
 \vdots & \vdots & \vdots & \vdots & \vdots \\
{w_{\rm{r}}^{N,1}}&{w_{\rm{r}}^{N,2}}& \cdots &{w_{\rm{r}}^{N,N}}&{w_{\rm{r}}^{N,N{\rm{ + }}1}}\\
{w_{\rm{r}}^{N{\rm{ + 1}},{\rm{1}}}}&{w_{\rm{r}}^{N{\rm{ + 1}},{\rm{2}}}}& \cdots &{w_{\rm{r}}^{N{\rm{ + }}1,N}}&{w_{\rm{r}}^{N{\rm{ + }}1,N{\rm{ + }}1}}
\end{array}} \right]\left[ {\begin{array}{*{20}{c}}
{{x_1}(n)}\\
{{x_2}(n)}\\
 \vdots \\
{{x_N}(n)}\\
0
\end{array}} \right]\\
{\rm{ = }}{\left[ {x_1^*\left( n \right),...,x_N^*\left( n \right),x_{N + 1}^*\left( n \right)} \right]^ \top },
\end{array}
\end{equation}   
\end{small}$x_{p}^{*}\left( n \right)\ne {{x}_{p}}\left( n \right)\left( p=1,2,...,N \right)$. The reservoir state of the initial nodes is altered, which clearly justifies the reason behind the special structure of the feedback matrix in Eq. (\ref{eq15}).
\end{remark}

\section{Convergence and stability}
In this section, we provide the stability and convergence analysis of the proposed approach based on the projection algorithm. Moreover, an enhanced condition is introduced to further improve the model's stability.
\subsection{Online learning of parameters}
Due to the dynamic changes in the actual industrial process, it is important and necessary to update the model parameters timely so that a better modelling performance can be achieved by using an updated RSCN model. Given an RSCN model, constructed through the offline learning, as follows:
\begin{equation}
\label{eq27}
{\bf{\hat x}}(n) = g({{\bf{\hat W}}_{{\rm{in}}}}{\bf{u}}(n) + {{\bf{\hat W}}_{\mathop{\rm r}\nolimits} }{\bf{\hat x}}(n - 1){\rm{ + }}{\bf{\hat b}}),
\end{equation}
\begin{equation}
\label{eq28}
{\bf{\hat y}}(n) = {{\bf{\hat W}}_{\rm{out}}}\left( {{\bf{\hat x}}(n),{\bf{u}}(n)} \right),
\end{equation}
where $\mathbf{\hat{x}}(n),\mathbf{\hat{y}}(n)$ are the predicted state vectors and output. ${{\mathbf{\hat{W}}}_{\text{in}}},{{\mathbf{\hat{W}}}_{\operatorname{r}}}$ are the input weight matrix and the feedback matrix, respectively, which are fixed. Denoted by $\mathbf{\hat{g}}(n)=\left( \mathbf{\hat{x}}(n),\mathbf{u}(n) \right)$, and a projection algorithm \cite{ref35} is applied to update the output weights ${{\mathbf{\hat{W}}}_{\rm{out}}}$.

Let ${\rm{H}} =\left\{ {{{\mathbf{\hat{W}}}}_{\rm{out}}}:\mathbf{\hat{y}}(n)={{{\mathbf{\hat{W}}}}_{\rm{out}}}\mathbf{\hat{g}}(n) \right\}$. Choose the closest weight to ${{\mathbf{\hat{W}}}_{\rm{out}}}\left( n-1 \right)$, and define ${{\mathbf{\hat{W}}}_{\rm{out}}}\left( n \right)$ through minimizing the following cost function:
\begin{equation}
\label{eq29}
\begin{array}{*{20}{c}}
{J = \frac{1}{2}{{\left\| {{{{\bf{\hat W}}}_{\rm{out}}}\left( n \right) - {{{\bf{\hat W}}}_{\rm{out}}}\left( {n - 1} \right)} \right\|}^2}}\\
{s.t.{\kern 1pt} {\kern 1pt} {\kern 1pt} {\kern 1pt} {\kern 1pt} {\kern 1pt} {\kern 1pt} {\kern 1pt} {\kern 1pt} {\kern 1pt} {\bf{y}}\left( n \right) = {{{\bf{\hat W}}}_{\rm{out}}}\left( n \right){\bf{\hat g}}(n)}
\end{array}.
\end{equation}
By introducing the Lagrange operator ${{\lambda }_{\text{p}}}$, we can obtain
\begin{equation}
\label{eq30}
\begin{array}{l}
{J_e} = \frac{1}{2}{\left\| {{{{\bf{\hat W}}}_{\rm{out}}}\left( n \right) - {{{\bf{\hat W}}}_{\rm{out}}}\left( {n - 1} \right)} \right\|^2}\\
{\kern 1pt} {\kern 1pt} {\kern 1pt} {\kern 1pt} {\kern 1pt} {\kern 1pt} {\kern 1pt} {\kern 1pt} {\kern 1pt} {\kern 1pt} {\kern 1pt} {\kern 1pt} {\kern 1pt} {\kern 1pt} {\kern 1pt} {\kern 1pt} {\kern 1pt} {\kern 1pt} {\kern 1pt} {\kern 1pt} {\kern 1pt} {\kern 1pt}  + {\lambda _{\rm{p}}}\left( {{\bf{y}}\left( n \right) - {{{\bf{\hat W}}}_{\rm{out}}}\left( n \right){\bf{\hat g}}(n)} \right).
\end{array}
\end{equation}
The necessary conditions for ${{J}_{e}}$ to be minimal are
\begin{equation}
\label{eq31}
\frac{{\partial {J_e}}}{{\partial {{{\bf{\hat W}}}_{\rm{out}}}\left( n \right)}} = 0{\kern 1pt} {\kern 1pt} {\kern 1pt} {\kern 1pt} {\kern 1pt} {\kern 1pt} {\kern 1pt} and{\kern 1pt} {\kern 1pt} {\kern 1pt} {\kern 1pt} {\kern 1pt} {\kern 1pt} \frac{{\partial {J_e}}}{{\partial {\lambda _{\rm{p}}}}} = 0{\kern 1pt} .
\end{equation}
Then, we have
\begin{equation}
\label{eq32}
\left\{ {\begin{array}{*{20}{c}}
{{{{\bf{\hat W}}}_{\rm{out}}}\left( n \right) - {{{\bf{\hat W}}}_{\rm{out}}}\left( {n - 1} \right) - {\lambda _{\rm{p}}}{{{\bf{\hat g}}}^ \top }(n) = 0}\\
{{\bf{y}}\left( n \right) - {{{\bf{\hat W}}}_{\rm{out}}}\left( n \right){\bf{\hat g}}(n) = 0}
\end{array}} \right.,
\end{equation}
\begin{equation}
\label{eq33}
{\lambda _{\rm{p}}} = \frac{{{\bf{y}}\left( n \right) - {{{\bf{\hat W}}}_{\rm{out}}}\left( {n - 1} \right){\bf{\hat g}}(n)}}{{{\bf{\hat g}}{{(n)}^ \top }{\bf{\hat g}}(n)}}.
\end{equation}
Substituting Eq. (\ref{eq33}) into Eq. (\ref{eq32}), the online update rule for the output weights can be formulated as
\begin{equation}
\label{eq34}
\begin{array}{l}
{{{\bf{\hat W}}}_{\rm{out}}}(n) = {{{\bf{\hat W}}}_{\rm{out}}}(n - 1)\\
{\kern 1pt}  + \frac{{{\bf{\hat g}}{{(n)}^ \top }}}{{{\bf{\hat g}}{{(n)}^ \top }{\bf{\hat g}}(n)}}\left( {{\bf{y}}\left( n \right) - {{{\bf{\hat W}}}_{\rm{out}}}\left( {n - 1} \right){\bf{\hat g}}(n)} \right).
\end{array}
\end{equation}
To avoid division by zero, a small constant $c$ is added to the denominator. Moreover, a coefficient $a$ can be multiplied by the numerator to obtain the improved projection algorithm
\begin{equation}
\label{eq35}
\begin{array}{l}
{{{\bf{\hat W}}}_{\rm{out}}}(n) = {{{\bf{\hat W}}}_{\rm{out}}}(n - 1)\\
{\kern 1pt}  + \frac{{a{\bf{\hat g}}{{(n)}^ \top }}}{{c + {\bf{\hat g}}{{(n)}^ \top }{\bf{\hat g}}(n)}}\left( {{\bf{y}}\left( n \right) - {{{\bf{\hat W}}}_{\rm{out}}}\left( {n - 1} \right){\bf{\hat g}}(n)} \right),
\end{array}
\end{equation}
where $0<a\le 1$ and $c>0$. 

\begin{algorithm}[t]\footnotesize 
\caption{Projection algorithm for updating output weights}
\KwIn{Given the input $\left[ \mathbf{u}(1),\mathbf{u}(2),\ldots ,\mathbf{u}(n) \right]$, the output $\left[ \mathbf{y}\left( 1 \right),\ldots ,\mathbf{y}\left( n \right) \right]$, initial state $\mathbf{\hat{x}}(0)$, offline trained input weight matrix ${{\mathbf{\hat{W}}}_{\text{in}}}$, feedback matrix ${{\mathbf{\hat{W}}}_{\operatorname{r}}}$, output weights ${{\mathbf{\hat{W}}}_{\text{out}}}$, and constants $a,c.$ } 
Initialization: Set the initial output weights as ${{\mathbf{\hat{W}}}_{\rm{out}}}(1)\text{=}{{\mathbf{\hat{W}}}_{\text{out}}};$ \\
Input the samples sequentially into Eq. (\ref{eq27}) and obtain $\left[ \mathbf{\hat{x}}(1),\ldots ,\mathbf{\hat{x}}(n) \right],$ $\left[ \mathbf{\hat{g}}(1),\ldots ,\mathbf{\hat{g}}(n) \right]$;\\
Calculate the output of the first step $\mathbf{\hat{y}}(1)={{\mathbf{\hat{W}}}_{\rm{out}}}\left( 1 \right)\mathbf{\hat{g}}(1)$;\\
\For{$i=2,3,\ldots ,n,$}{
    Update ${{\mathbf{\hat{W}}}_{\rm{out}}}(i)$ according to Eq. (\ref{eq35});\\
    Calculate $\mathbf{\hat{y}}(i)={{\mathbf{\hat{W}}}_{\rm{out}}}\left( i \right)\mathbf{\hat{g}}(i)$; 
    }
 \textbf{Return:} $\left[ {{{\mathbf{\hat{W}}}}_{\rm{out}}}\left( 1 \right),\ldots ,{{{\mathbf{\hat{W}}}}_{\rm{out}}}\left( n \right) \right]$.
\end{algorithm} 

\subsection{Convergence analysis}
Convergence is one of the key performances in data modelling, which tells if the algorithm can achieve a stable solution over time. In this subsection, we give a convergence analysis on the output weights as follows. 

Subtracting ${{\mathbf{\hat{W}}}_{0}}$ on both the sides of Eq. (\ref{eq35}), we have
\begin{equation}
\label{eq43}
\begin{array}{l}
{{{\bf{\hat W}}}_{\rm{out}}}(n) - {{{\bf{\hat W}}}_0}\\
 = \left( {{{{\bf{\hat W}}}_{\rm{out}}}(n - 1) - {{{\bf{\hat W}}}_0}} \right)\\
{\kern 1pt} {\kern 1pt} {\kern 1pt} {\kern 1pt} {\kern 1pt} {\kern 1pt} {\kern 1pt} {\kern 1pt}  + \frac{{a{\bf{\hat g}}{{(n)}^ \top }}}{{c + {\bf{\hat g}}{{(n)}^ \top }{\bf{\hat g}}(n)}}\left( {{\bf{y}}\left( n \right) - {{{\bf{\hat W}}}_{\rm{out}}}\left( {n - 1} \right){\bf{\hat g}}(n)} \right)\\
 = \left( {{{{\bf{\hat W}}}_{\rm{out}}}(n - 1) - {{{\bf{\hat W}}}_0}} \right)\\
{\kern 1pt} {\kern 1pt} {\kern 1pt} {\kern 1pt} {\kern 1pt} {\kern 1pt} {\kern 1pt} {\kern 1pt} {\kern 1pt} {\kern 1pt} {\kern 1pt} {\kern 1pt} {\kern 1pt} {\kern 1pt}  - \frac{{a\left( {{{{\bf{\hat W}}}_{\rm{out}}}(n - 1) - {{{\bf{\hat W}}}_0}} \right){\bf{\hat g}}{{(n)}^ \top }{\bf{\hat g}}(n)}}{{c + {\bf{\hat g}}{{(n)}^ \top }{\bf{\hat g}}(n)}}\\
 = \left( {{{{\bf{\hat W}}}_{\rm{out}}}(n - 1) - {{{\bf{\hat W}}}_0}} \right)\hat \lambda \left( n \right),
\end{array}
\end{equation}
where $\hat{\lambda }(n)=1-\frac{a\mathbf{\hat{g}}{{(n)}^{\top }}\mathbf{\hat{g}}(n)}{c+\mathbf{\hat{g}}{{(n)}^{\top }}\mathbf{\hat{g}}(n)}$, ${{\mathbf{\hat{W}}}_{0}}$ is an ideal output weight satisfying $\mathbf{y}\left( n \right)\approx {{\mathbf{\hat{W}}}_{0}}\mathbf{\hat{g}}(n)$. Obviously, $\left| {\hat \lambda (n)} \right| \le 1$, thus we have
\begin{small}
 \begin{equation}
\label{eq44}
\left| {{{{\bf{\hat W}}}_{\rm{out}}}(n) - {{{\bf{\hat W}}}_0}} \right| \le  \ldots \le\left| {{{{\bf{\hat W}}}_{\rm{out}}}(0) - {{{\bf{\hat W}}}_0}} \right|.
\end{equation}   
\end{small}

From Eq. (\ref{eq44}), we cannot conclude that ${{\mathbf{\hat{W}}}_{\rm{out}}}(n)$ must converge to ${{\mathbf{\hat{W}}}_{0}}$. It only guarantees that ${{\mathbf{\hat{W}}}_{\rm{out}}}(n)$ does not deviate further from difference between ${{\mathbf{\hat{W}}}_{0}}$ than ${{\mathbf{\hat{W}}}_{\rm{out}}}(0)$. By referring to \cite{ref35}, we can establish the following result. \\

\textbf{Theorem 3.} With the following updating rule for the output weight ${{\mathbf{\hat{W}}}_{\rm{out}}}(n)$,
 \begin{equation}
\label{eq45}
\begin{array}{l}
{{{\bf{\hat W}}}_{\rm{out}}}(n) = {{{\bf{\hat W}}}_{\rm{out}}}(n - 1)\\
{\kern 1pt}  + \frac{{{\bf{\hat g}}{{(n)}^ \top }}}{{\sum\limits_{l = 1}^n {{\bf{\hat g}}{{(n)}^ \top }{\bf{\hat g}}(n)} }}\left( {{\bf{y}}\left( n \right) - {{{\bf{\hat W}}}_{\rm{out}}}\left( {n - 1} \right){\bf{\hat g}}(n)} \right),
\end{array}
\end{equation} 
we have 
 \begin{equation}
\label{eq451}
\underset{n\to \infty }{\mathop{\lim }}\,{{\mathbf{\hat{W}}}_{\rm{out}}}(n)={{\mathbf{\hat{W}}}_{0}}.
\end{equation} 

\textbf{Proof.} At time step $n$, the approximation error can be expressed as
\begin{equation}
\label{eq46}
{e_{\rm{m}}}\left( n \right) = {{\bf{\hat W}}_0}{\bf{\hat g}}(n) - {{\bf{\hat W}}_{\rm{out}}}(n){\bf{\hat g}}(n){\rm{ = }}\left( {{{{\bf{\hat W}}}_0} - {{{\bf{\hat W}}}_{\rm{out}}}(n)} \right){\bf{\hat g}}(n),
\end{equation}
\begin{equation}
\label{eq47}
\begin{array}{l}
{e_{\rm{s}}}\left( n \right) = {{{\bf{\hat W}}}_0}{\bf{\hat g}}(n) - {{{\bf{\hat W}}}_{\rm{out}}}(n - 1){\bf{\hat g}}(n)\\
{\kern 1pt} {\kern 1pt} {\kern 1pt} {\kern 1pt} {\kern 1pt} {\kern 1pt} {\kern 1pt} {\kern 1pt} {\kern 1pt} {\kern 1pt} {\kern 1pt} {\kern 1pt} {\kern 1pt} {\kern 1pt} {\kern 1pt} {\kern 1pt} {\kern 1pt} {\kern 1pt} {\kern 1pt} {\kern 1pt} {\kern 1pt} {\kern 1pt} {\kern 1pt} {\kern 1pt} {\kern 1pt} {\kern 1pt} {\kern 1pt} {\rm{ = }}\left( {{{{\bf{\hat W}}}_0} - {{{\bf{\hat W}}}_{\rm{out}}}(n - 1)} \right){\bf{\hat g}}(n).
\end{array}
\end{equation}
Let \small${{\mathbf{\hat{W}}}_{\Delta }}(n)={{\mathbf{\hat{W}}}_{\rm{out}}}(n)-{{\mathbf{\hat{W}}}_{0}}$, \normalsize and \small$\mathbf{H}\left( n \right)={{\left( \sum\limits_{l=1}^{n}{\mathbf{\hat{g}}{{(l)}^{\top }}\mathbf{\hat{g}}(l)} \right)}^{-1}}$. \normalsize Combining Eq. (\ref{eq45}) and Eq. (\ref{eq47}), we have 
\begin{equation}
\label{eq48}
\begin{array}{l}
{{{\bf{\hat W}}}_{\rm{out}}}(n) = {{{\bf{\hat W}}}_\Delta }(n)-{{{\bf{\hat W}}}_0}\\
{\kern 1pt}  = {{{\bf{\hat W}}}_{\rm{out}}}(n - 1) + {\bf{H}}\left( n \right){\bf{\hat g}}{(n)^ \top }{e_s}\left( n \right)\\
{\kern 1pt}  ={{{\bf{\hat W}}}_\Delta }(n - 1)-{{{\bf{\hat W}}}_0} - {\bf{H}}\left( n \right){\bf{\hat g}}{(n)^ \top }{\bf{\hat g}}(n){{{\bf{\hat W}}}_\Delta }(n - 1),
\end{array}
\end{equation}
\begin{equation}
\label{eq49}
{{\bf{\hat W}}_\Delta }(n) = \left( {{\bf{I}} - {\bf{H}}\left( n \right){\bf{\hat g}}{{(n)}^ \top }{\bf{\hat g}}(n)} \right){{\bf{\hat W}}_\Delta }(n - 1),
\end{equation}
where $\mathbf{I}$ is the identity matrix. Substituting ${{\bf{H}}^{ - 1}}\left( n \right) = {{\bf{H}}^{ - 1}}\left( {n - 1} \right) + {\bf{\hat g}}{(n)^ \top }{\bf{\hat g}}(n)$ into Eq. (\ref{eq49}), we obtain
\begin{equation}
\label{eq50}
{{\bf{\hat W}}_\Delta }(n) = {\bf{H}}\left( n \right){{\bf{H}}^{ - 1}}\left( {n - 1} \right){{\bf{\hat W}}_\Delta }(n - 1).
\end{equation}
Let $\mathbf{P}\left( n \right)=\mathbf{H}\left( n \right){{\mathbf{H}}^{-1}}\left( n-1 \right)$. Eq. (\ref{eq50}) can be rewritten as
\begin{equation}
\label{eq51}
{{\bf{\hat W}}_\Delta }(n) = {\bf{P}}\left( n \right){{\bf{\hat W}}_\Delta }(n - 1).
\end{equation}
Furthermore, the matrix $\mathbf{P}\left( n \right)$ can be reconstructed by
\begin{small}
 \begin{equation}
\label{eq52}
{\bf{P}}\left( n \right) = {\left( {{{\bf{H}}^{ - 1}}\left( {n - 1} \right) + {\bf{\hat g}}{{(n)}^ \top }{\bf{\hat g}}(n)} \right)^{ - 1}}{{\bf{H}}^{ - 1}}\left( {n - 1} \right),
\end{equation}   
\end{small}where ${{\bf{H}}^{ - 1}}\left( {n - 1} \right) = \sum\limits_{l = 1}^{n - 1} {{\bf{\hat g}}{{(l)}^ \top }{\bf{\hat g}}(l)}$ is a symmetric and positive matrix, and so is ${\bf{\hat g}}{(n)^ \top }{\bf{\hat g}}(n)$. Therefore, the norm of the matrix $\mathbf{P}\left( n \right)$ is less than 1, that is, $\left\| {{\bf{P}}\left( n \right)} \right\| \le 1 - {\varepsilon _{\rm{o}}}$, where ${\varepsilon _{\rm{o}}} > 0$ is a very small number. It can be easily shown that 
\begin{equation}
\label{eq53}
\left\| {{{{\bf{\hat W}}}_\Delta }(n)} \right\| < {\left( {1 - {\varepsilon _{\rm{o}}}} \right)^n}\left\| {{{{\bf{\hat W}}}_\Delta }(0)} \right\|.
\end{equation}
Thus, we have $\underset{n\to \infty }{\mathop{\lim }}\,{{\mathbf{\hat{W}}}_{\Delta }}(n)=0$, that is $\underset{n\to \infty }{\mathop{\lim }}\,{{\mathbf{\hat{W}}}_{\rm{out}}}(n)={{\mathbf{\hat{W}}}_{0}}$, which completes the proof.
\subsection{Stability analysis}
This subsection presents a theoretical analysis on the stability of the modelling performance.\\

\textbf{Theorem 4.} Let $\mathbf{y}\left( n \right)={{\mathbf{\hat{W}}}_{0}}\mathbf{\hat{g}}(n)+\nu \left( n \right)$, where $\nu \left( n \right)$ is a compensation term, and $\sup \left[ {\nu \left( n \right)} \right] = \varphi \left( n \right)$. Define
${{\tilde{e}}_{\text{m}}}\left( n \right)=\left( {{{\mathbf{\hat{W}}}}_{0}}-{{{\mathbf{\hat{W}}}}_{\text{out}}}(n) \right)\mathbf{\hat{g}}(n)+\nu (n)={{\mathbf{\hat{W}}}_{\Delta }}(n)\mathbf{\hat{g}}(n)+\nu (n)$. Substitute the coefficient $a$ in Eq. (\ref{eq43}) with $\eta \left( n \right)$, and set $c=1$, where 
\begin{equation}
\label{eq531}
\eta \left( n \right) = \left\{ {\begin{array}{*{20}{c}}
{1,{\kern 1pt} {\kern 1pt} {\kern 1pt} {\kern 1pt} {\kern 1pt} {\kern 1pt} {\kern 1pt} {\kern 1pt} {\kern 1pt} {\kern 1pt} {\kern 1pt} \left| {{{\tilde e}_{\rm{m}}}\left( n \right)} \right| > 2\varphi \left( n \right)}\\
{0,{\kern 1pt} {\kern 1pt} {\kern 1pt} {\kern 1pt} {\kern 1pt} {\kern 1pt} {\kern 1pt} {\kern 1pt} {\kern 1pt} {\kern 1pt} {\kern 1pt} {\kern 1pt}{\kern 1pt} {\kern 1pt} {\kern 1pt} {\kern 1pt} {\kern 1pt} {\kern 1pt} {\kern 1pt} {\kern 1pt} otherwise{\kern 1pt} {\kern 1pt} {\kern 1pt} {\kern 1pt} {\kern 1pt} {\kern 1pt} {\kern 1pt} {\kern 1pt} {\kern 1pt} {\kern 1pt} {\kern 1pt} {\kern 1pt} {\kern 1pt} {\kern 1pt} {\kern 1pt} {\kern 1pt} {\kern 1pt} {\kern 1pt} {\kern 1pt} {\kern 1pt} {\kern 1pt} }
\end{array}} \right..
\end{equation}
Then, the following results can be established, that is, 
\begin{small}
\begin{equation}
\label{eq54}
\mathop {\lim }\limits_{M \to \infty } \sum\limits_{n = 1}^M {\frac{{\eta (n)\left( {\tilde e_{\rm{m}}^2\left( n \right) - 4{\nu ^2}\left( n \right)} \right)}}{{2\left( {1 + {\bf{\hat g}}{{(n)}^ \top }{\bf{\hat g}}(n)} \right)}} < \infty },
\end{equation}
\end{small}and
\begin{small}
\begin{equation}
\label{eq55}
\mathop {\lim }\limits_{n \to \infty } \frac{{\eta (n)\left( {\tilde e_{\rm{m}}^2\left( n \right) - 4{\nu ^2}\left( n \right)} \right)}}{{2\left( {1 + {\bf{\hat g}}{{(n)}^ \top }{\bf{\hat g}}(n)} \right)}} = 0.
\end{equation}
\end{small}

\textbf{Proof.} With simple computation, we have
\begin{equation}
\label{eq56}
{{\bf{\hat W}}_\Delta }(n) = {{\bf{\hat W}}_\Delta }(n - 1) + \frac{{\eta (n){\bf{\hat g}}{{(n)}^ \top }{{\tilde e}_{\rm{m}}}\left( n \right)}}{{1 + {\bf{\hat g}}{{(n)}^ \top }{\bf{\hat g}}(n)}},
\end{equation}
and
\begin{equation}
\label{eq57}
\begin{array}{l}
{\left\| {{{{\bf{\hat W}}}_\Delta }(n)} \right\|^2}\\
 = {\left\| {{{{\bf{\hat W}}}_\Delta }(n - 1)} \right\|^2} + \frac{{2\eta (n){{{\bf{\hat W}}}_\Delta }(n - 1){\bf{\hat g}}{{(n)}^ \top }{{\tilde e}_{\rm{m}}}\left( n \right)}}{{1 + {\bf{\hat g}}{{(n)}^ \top }{\bf{\hat g}}(n)}}\\
{\kern 1pt} {\kern 1pt} {\kern 1pt} {\kern 1pt} {\kern 1pt} {\kern 1pt} {\kern 1pt} {\kern 1pt} {\kern 1pt} {\kern 1pt} {\kern 1pt} {\kern 1pt}  + \frac{{{\eta ^2}(n)\tilde e_{\rm{m}}^2\left( n \right){{\left\| {{\bf{\hat g}}(n)} \right\|}^2}}}{{{{\left( {1 + {\bf{\hat g}}{{(n)}^ \top }{\bf{\hat g}}(n)} \right)}^2}}}\\
 = {\left\| {{{{\bf{\hat W}}}_\Delta }(n - 1)} \right\|^2} + \frac{{2\eta (n){{\tilde e}_{\rm{m}}}\left( n \right)\nu \left( n \right)}}{{1 + {\bf{\hat g}}{{(n)}^ \top }{\bf{\hat g}}(n)}}\\
 - \frac{{\eta (n)\tilde e_{\rm{m}}^2\left( n \right)}}{{1 + {\bf{\hat g}}{{(n)}^ \top }{\bf{\hat g}}(n)}}\left( {2 + \frac{{\eta (n){{\left\| {{\bf{\hat g}}(n)} \right\|}^2}}}{{1 + {\bf{\hat g}}{{(n)}^ \top }{\bf{\hat g}}(n)}}} \right).
\end{array}
\end{equation}
Due to $2{{C}_{1}}{{C}_{2}}\le DC_{1}^{2}+{C_{2}^{2}}/{D}\;$ for any positive $D$, yields
\begin{equation}
\label{eq58}
\begin{array}{l}
{\left\| {{{{\bf{\hat W}}}_\Delta }(n)} \right\|^2} \le {\left\| {{{{\bf{\hat W}}}_\Delta }(n - 1)} \right\|^2} + \frac{{2\eta (n){{\tilde e}_{\rm{m}}}\left( n \right)\nu \left( n \right)}}{{1 + {\bf{\hat g}}{{(n)}^ \top }{\bf{\hat g}}(n)}}\\
{\kern 1pt} {\kern 1pt} {\kern 1pt} {\kern 1pt} {\kern 1pt} {\kern 1pt} {\kern 1pt} {\kern 1pt} {\kern 1pt} {\kern 1pt} {\kern 1pt} {\kern 1pt} {\kern 1pt} {\kern 1pt} {\kern 1pt} {\kern 1pt} {\kern 1pt} {\kern 1pt} {\kern 1pt} {\kern 1pt} {\kern 1pt} {\kern 1pt} {\kern 1pt} {\kern 1pt} {\kern 1pt} {\kern 1pt} {\kern 1pt} {\kern 1pt} {\kern 1pt} {\kern 1pt} {\kern 1pt} {\kern 1pt} {\kern 1pt} {\kern 1pt} {\kern 1pt} {\kern 1pt} {\kern 1pt} {\kern 1pt} {\kern 1pt} {\kern 1pt} {\kern 1pt} {\kern 1pt} {\kern 1pt} {\kern 1pt} {\kern 1pt} {\kern 1pt} {\kern 1pt}  - \frac{{\eta (n)\tilde e_{\rm{m}}^2\left( n \right)}}{{1 + {\bf{\hat g}}{{(n)}^ \top }{\bf{\hat g}}(n)}}\\
{\kern 1pt} {\kern 1pt} {\kern 1pt} {\kern 1pt} {\kern 1pt} {\kern 1pt} {\kern 1pt} {\kern 1pt} {\kern 1pt} {\kern 1pt} {\kern 1pt} {\kern 1pt} {\kern 1pt} {\kern 1pt} {\kern 1pt} {\kern 1pt} {\kern 1pt} {\kern 1pt} {\kern 1pt} {\kern 1pt} {\kern 1pt} {\kern 1pt} {\kern 1pt} {\kern 1pt} {\kern 1pt} {\kern 1pt} {\kern 1pt} {\kern 1pt} {\kern 1pt} {\kern 1pt} {\kern 1pt} {\kern 1pt} {\kern 1pt} {\kern 1pt} {\kern 1pt} {\kern 1pt} {\kern 1pt} {\kern 1pt} {\kern 1pt} {\kern 1pt} {\kern 1pt} {\kern 1pt} {\kern 1pt} {\kern 1pt} {\kern 1pt} {\kern 1pt} {\kern 1pt}  \le {\left\| {{{{\bf{\hat W}}}_\Delta }(n - 1)} \right\|^2} + \frac{{\eta (n)\left( {{\raise0.7ex\hbox{${\tilde e_{\rm{m}}^2\left( n \right)}$} \!\mathord{\left/
 {\vphantom {{\tilde e_{\rm{m}}^2\left( n \right)} 2}}\right.\kern-\nulldelimiterspace}
\!\lower0.7ex\hbox{$2$}} + 2{\nu ^2}\left( n \right)} \right)}}{{1 + {\bf{\hat g}}{{(n)}^ \top }{\bf{\hat g}}(n)}}\\
{\kern 1pt} {\kern 1pt} {\kern 1pt} {\kern 1pt} {\kern 1pt} {\kern 1pt} {\kern 1pt} {\kern 1pt} {\kern 1pt} {\kern 1pt} {\kern 1pt} {\kern 1pt} {\kern 1pt} {\kern 1pt} {\kern 1pt} {\kern 1pt} {\kern 1pt} {\kern 1pt} {\kern 1pt} {\kern 1pt} {\kern 1pt} {\kern 1pt} {\kern 1pt} {\kern 1pt} {\kern 1pt} {\kern 1pt} {\kern 1pt} {\kern 1pt} {\kern 1pt} {\kern 1pt} {\kern 1pt} {\kern 1pt} {\kern 1pt} {\kern 1pt} {\kern 1pt} {\kern 1pt} {\kern 1pt} {\kern 1pt} {\kern 1pt} {\kern 1pt} {\kern 1pt} {\kern 1pt} {\kern 1pt} {\kern 1pt} {\kern 1pt} {\kern 1pt} {\kern 1pt} {\kern 1pt}  - \frac{{\eta (n)\tilde e_{\rm{m}}^2\left( n \right)}}{{1 + {\bf{\hat g}}{{(n)}^ \top }{\bf{\hat g}}(n)}}\\
{\kern 1pt} {\kern 1pt} {\kern 1pt} {\kern 1pt} {\kern 1pt} {\kern 1pt} {\kern 1pt} {\kern 1pt} {\kern 1pt} {\kern 1pt} {\kern 1pt} {\kern 1pt} {\kern 1pt} {\kern 1pt} {\kern 1pt} {\kern 1pt} {\kern 1pt} {\kern 1pt} {\kern 1pt} {\kern 1pt} {\kern 1pt} {\kern 1pt} {\kern 1pt} {\kern 1pt} {\kern 1pt} {\kern 1pt} {\kern 1pt} {\kern 1pt} {\kern 1pt} {\kern 1pt} {\kern 1pt} {\kern 1pt} {\kern 1pt} {\kern 1pt} {\kern 1pt} {\kern 1pt} {\kern 1pt} {\kern 1pt} {\kern 1pt} {\kern 1pt} {\kern 1pt} {\kern 1pt} {\kern 1pt} {\kern 1pt} {\kern 1pt} {\kern 1pt}  \le {\left\| {{{{\bf{\hat W}}}_\Delta }(n - 1)} \right\|^2} - \frac{{\eta (n)\left( {\tilde e_{\rm{m}}^2\left( n \right) - 4{\nu ^2}\left( n \right)} \right)}}{{2\left( {1 + {\bf{\hat g}}{{(n)}^ \top }{\bf{\hat g}}(n)} \right)}}.
\end{array}
\end{equation}
From Eq. (\ref{eq44}) or Eq. (\ref{eq53}), we know that
\begin{equation}
\label{eq59}
{\left\| {{{{\bf{\hat W}}}_\Delta }(n)} \right\|^2} \le {\left\| {{{{\bf{\hat W}}}_\Delta }(n - 1)} \right\|^2} \le {\left\| {{{{\bf{\hat W}}}_\Delta }(0)} \right\|^2},
\end{equation}
Therefore, we have
\begin{footnotesize}
\begin{equation}
\label{eq60}
\mathop {\lim }\limits_{M \to \infty } \sum\limits_{n = 1}^M {\frac{{\eta (n)\left( {\tilde e_{\rm{m}}^2\left( n \right) - 4{\nu ^2}\left( n \right)} \right)}}{{2\left( {1 + {\bf{\hat g}}{{(n)}^ \top }{\bf{\hat g}}(n)} \right)}} \le \left\| {{{{\bf{\hat W}}}_\Delta }(0)} \right\| - \left\| {{{{\bf{\hat W}}}_\Delta }(n)} \right\|},
\end{equation}
\end{footnotesize}which implies $\mathop {\lim }\limits_{n \to \infty } \frac{{\eta (n)\left( {\tilde e_{\rm{m}}^2\left( n \right) - 4{\nu ^2}\left( n \right)} \right)}}{{2\left( {1 + {\bf{\hat g}}{{(n)}^ \top }{\bf{\hat g}}(n)} \right)}} = 0$. This completes the proof.

Furthermore, we can infer that \footnotesize $\mathop {\lim }\limits_{n \to \infty } \eta (n)\left( {\tilde e_{\rm{m}}^2\left( n \right) - 4{\nu ^2}\left( n \right)} \right) = 0$. \normalsize According to the definition of ${\eta (n)}$, we have $\mathop {\lim }\limits_{n \to \infty } \sup \left[ {\tilde e_{\rm{m}}^2\left( n \right)} \right] \le \sup \left[ {4{\nu ^2}\left( n \right)} \right]$. Therefore, following inequality can be established, that is,
\begin{equation}
\label{eq63}
\mathop {\lim }\limits_{n \to \infty } \sup \left[ {\tilde e_{\rm{m}}^{}\left( n \right)} \right] \le \sup \left[ {2\nu \left( n \right)} \right] = 2\varphi \left( n \right).
\end{equation}
Based on Eq. (\ref{eq54})-Eq. (\ref{eq63}), for $\forall {{\varepsilon }_{\text{m}}}>0$, $\exists M$, as $n>M$,  
\begin{equation}
\label{eq64}
\left| {\tilde e_{\rm{m}}^{}\left( n \right)} \right| < 2\varphi \left( n \right) + {\varepsilon _{\rm{m}}}.
\end{equation}
where ${{\varepsilon }_{\text{m}}}$ is a very small number. Since the RSCN is a universal approximator, it can approximate any function on a compact set with arbitrary accuracy. Thus, we can obtain that $\varphi \left( n \right)\to 0$ and
$\left| \tilde{e}_{\text{m}}^{{}}\left( n \right) \right|<{{{\varepsilon }'}_{\text{m}}},$
where ${{{\varepsilon }'}_{\text{m}}}=2\varphi \left( n \right)+{{\varepsilon }_{\text{m}}}$. This demonstrates that the modelling error can converge to a smaller neighborhood of the origin.

In addition, a stronger result on the modelling performance stability can be established under some assumptions. \\

\textbf{Theorem 5.} Let $\Delta \mathbf{\hat{g}}(n-1)=\mathbf{\hat{g}}(n)-\mathbf{\hat{g}}(n-1)$, ${{\eta }_{a}}={{e}_{\text{m}}}\left( n-1 \right)$, ${{\eta }_{b}}=\left( {{{\mathbf{\hat{W}}}}_{0}}-{{{\mathbf{\hat{W}}}}_{\rm{out}}}(n-1) \right)\Delta \mathbf{\hat{g}}(n-1)$. The resulting model holds the universal approximation property, that is $\mathop {\lim }\limits_{n \to \infty } {{e}_{\text{m}}}\left( n \right) = 0$, if
\begin{equation}
\label{eq65}
{\eta _a}{\eta _b} < 0{\kern 1pt} {\kern 1pt} {\kern 1pt} {\kern 1pt} {\kern 1pt} {\kern 1pt} {\kern 1pt} and{\kern 1pt} {\kern 1pt} {\kern 1pt} {\kern 1pt} {\kern 1pt} {\kern 1pt} \left| {{\eta _b}} \right| \le 2\left| {{\eta _a}} \right|.
\end{equation}

\textbf{Proof.} According to Eq. (\ref{eq43}), $\left| {{e}_{\text{m}}}\left( n \right) \right|$ can be written as
\begin{equation}
\label{eq66}
\begin{array}{l}
\left| {{e_{\rm{m}}}\left( n \right)} \right| = \left| {\left( {{{{\bf{\hat W}}}_0} - {{{\bf{\hat W}}}_{\rm{out}}}(n)} \right){\bf{\hat g}}(n)} \right|\\
 = \left| {\left( {1 - \frac{{a{\bf{\hat g}}{{(n)}^ \top }{\bf{\hat g}}(n)}}{{c + {\bf{\hat g}}{{(n)}^ \top }{\bf{\hat g}}(n)}}} \right)\left( {{{{\bf{\hat W}}}_0} - {{{\bf{\hat W}}}_{\rm{out}}}(n - 1)} \right){\bf{\hat g}}(n)} \right|\\
 = \hat \lambda (n)\left| {\left( {{{{\bf{\hat W}}}_0} - {{{\bf{\hat W}}}_{\rm{out}}}(n - 1)} \right){\bf{\hat g}}(n)} \right|\\
= \hat \lambda (n)\left| {{e_{\rm{s}}}\left( n \right)} \right|.
\end{array}
\end{equation}
Let ${\hat \lambda _{\max }}$ denote the maximum value of $\hat \lambda (n)$. For $0<a\le 1$, $c>0$, it is easy to verify that $0 < \hat \lambda (n) \le {\hat \lambda _{\max }} < 1$, so $\left| {{e}_{\text{m}}}\left( n \right) \right|< \left| {e_{\rm{s}}}\left( n \right) \right|.$ For ${e_{\rm{s}}}\left( n \right)$, we can rewrite it as
\begin{equation}
\label{eq67}
\begin{array}{l}
{e_{\rm{s}}}\left( n \right) = {{{\bf{\hat W}}}_0}{\bf{\hat g}}(n) - {{{\bf{\hat W}}}_{\rm{out}}}(n - 1){\bf{\hat g}}(n)\\
{\kern 1pt} {\kern 1pt} {\kern 1pt} {\kern 1pt} {\kern 1pt} {\kern 1pt} {\kern 1pt} {\kern 1pt} {\kern 1pt} {\kern 1pt} {\kern 1pt} {\kern 1pt} {\kern 1pt} {\kern 1pt} {\kern 1pt} {\kern 1pt} {\kern 1pt} {\kern 1pt} {\kern 1pt} {\kern 1pt} {\kern 1pt} {\kern 1pt} {\kern 1pt} {\kern 1pt} {\kern 1pt} {\kern 1pt} {\kern 1pt} {\kern 1pt} {\rm{ = }}{{{\bf{\hat W}}}_0}\left( {{\bf{\hat g}}(n - 1) + \Delta {\bf{\hat g}}(n - 1)} \right)\\
{\kern 1pt} {\kern 1pt} {\kern 1pt} {\kern 1pt} {\kern 1pt} {\kern 1pt} {\kern 1pt} {\kern 1pt} {\kern 1pt} {\kern 1pt} {\kern 1pt} {\kern 1pt} {\kern 1pt} {\kern 1pt} {\kern 1pt} {\kern 1pt} {\kern 1pt} {\kern 1pt} {\kern 1pt} {\kern 1pt} {\kern 1pt} {\kern 1pt} {\kern 1pt} {\kern 1pt} {\kern 1pt} {\kern 1pt} {\kern 1pt} {\kern 1pt} {\kern 1pt}  - {{{\bf{\hat W}}}_{\rm{out}}}(n - 1)\left( {{\bf{\hat g}}(n - 1) + \Delta {\bf{\hat g}}(n - 1)} \right)\\
{\kern 1pt} {\kern 1pt} {\kern 1pt} {\kern 1pt} {\kern 1pt} {\kern 1pt} {\kern 1pt} {\kern 1pt} {\kern 1pt} {\kern 1pt} {\kern 1pt} {\kern 1pt} {\kern 1pt} {\kern 1pt} {\kern 1pt} {\kern 1pt} {\kern 1pt} {\kern 1pt} {\kern 1pt} {\kern 1pt} {\kern 1pt} {\kern 1pt} {\kern 1pt} {\kern 1pt} {\kern 1pt} {\kern 1pt}  = \left( {{{{\bf{\hat W}}}_0} - {{{\bf{\hat W}}}_{\rm{out}}}(n - 1)} \right){\bf{\hat g}}(n - 1)\\
{\kern 1pt} {\kern 1pt} {\kern 1pt} {\kern 1pt} {\kern 1pt} {\kern 1pt} {\kern 1pt} {\kern 1pt} {\kern 1pt} {\kern 1pt} {\kern 1pt} {\kern 1pt} {\kern 1pt} {\kern 1pt} {\kern 1pt} {\kern 1pt} {\kern 1pt} {\kern 1pt} {\kern 1pt} {\kern 1pt} {\kern 1pt} {\kern 1pt} {\kern 1pt} {\kern 1pt} {\kern 1pt} {\kern 1pt} {\kern 1pt} {\kern 1pt} {\kern 1pt} {\kern 1pt}  + \left( {{{{\bf{\hat W}}}_0} - {{{\bf{\hat W}}}_{\rm{out}}}(n - 1)} \right)\Delta {\bf{\hat g}}(n - 1)\\
 {\kern 1pt} {\kern 1pt} {\kern 1pt} {\kern 1pt} {\kern 1pt} {\kern 1pt} {\kern 1pt} {\kern 1pt} {\kern 1pt} {\kern 1pt} {\kern 1pt} {\kern 1pt} {\kern 1pt} {\kern 1pt} {\kern 1pt} {\kern 1pt} {\kern 1pt} {\kern 1pt} {\kern 1pt} {\kern 1pt} {\kern 1pt} {\kern 1pt} {\kern 1pt} {\kern 1pt} {\kern 1pt} {\kern 1pt}  = {\eta _a} + {\eta _b}.
\end{array}
\end{equation}
Note that ${{\eta }_{a}}$ and ${{\eta }_{b}}$ satisfy the constraints in Eq. (\ref{eq65}), that is,
\begin{equation}
\label{eq68}
\left\{ {\begin{array}{*{20}{c}}
{ - 2{\eta _a} \le {\eta _b} < 0,{\kern 1pt} {\kern 1pt} {\kern 1pt} {\kern 1pt} {\kern 1pt} {\kern 1pt} {\kern 1pt} {\kern 1pt} {\eta _a} > 0}\\
{0 < {\eta _b} \le  - 2{\eta _a},{\kern 1pt} {\kern 1pt} {\kern 1pt} {\kern 1pt} {\kern 1pt} {\kern 1pt} {\kern 1pt} {\kern 1pt} {\eta _a} < 0}
\end{array}} \right..
\end{equation}
Thus, we have $\left| {{\eta }_{a}}+{{\eta }_{b}} \right|\le \left| {{\eta }_{a}} \right|$, which implies that $\left| {e_{\rm{s}}}\left( n \right) \right|\le \left| {{e}_{\text{m}}}\left( n-1 \right) \right|$. Combining Eq. (\ref{eq43}) and Eq. (\ref{eq66}), we have $\left| {{e_{\rm{m}}}\left( n \right)} \right| \le {\hat \lambda _{\max }}\left| {{e_{\rm{s}}}\left( n \right)} \right| \le {\hat \lambda _{\max }}\left| {{e_{\rm{m}}}\left( {n - 1} \right)} \right|$, which completes the proof.\\

\section{Experimental results}
In this section, the performance of the proposed RSCNs is evaluated on four modelling tasks: Mackey-Glass time-series prediction, a nonlinear identification problem, and two industry data predictive analyses. The experimental results are compared with those obtained from the LSTM network, the original ESN, and several state-of-the-art ESN methods, including the SCR \cite{ref19}, leaky-integrator ESN (LIESN) \cite{ref21}, and polynomial ESN (PESN) \cite{ref34}.

\subsection{Evaluation metric and parameter settings}
In our experiments, the normalized root means square error (NRMSE) is used to evaluate the model performance, that is,
\begin{equation}
\label{eq69}
NRMSE = \sqrt {\frac{{\sum\limits_{n = 1}^{{n_{max}}} {{{\left( {{\bf{y}}\left( n \right) - {\bf{t}}\left( n \right)} \right)}^2}} }}{{{n_{max}}{\mathop{\rm var}} \left( {\bf{t}} \right)}}} ,
\end{equation}
where $\operatorname{var}\left( \mathbf{t} \right)$ is the variance of the desired output $\mathbf{t}$. 

The key parameters are taken as follows: the scaling factor of spectral radius varies from 0.5 to 1, and the sparsity of the feedback matrix is set to 0.03. The scope setting of input weight and feedback matrix for the ESN, SCR, PESN, and LIESN is $\lambda =1$. For LSTM, the number of iterations is set to 1000. And the weights and biases are initialized in $\left[ -1,1 \right]$. The online learning rate varies from 0.01 to 0.1. The error tolerance is set as ${{\varepsilon }_{LSTM}}={{10}^{-6}}$. RSCN is set with the following parameters: the maximum number of stochastic configurations ${{G}_{\max }}=100$, the weights scale sequence $\left\{ 0.5,1,5,10,30,50,100 \right\}$, the contractive sequence $r=\left[ 0.9,0.99,0.999,0.9999,0.99999 \right]$, the training tolerance ${{\varepsilon }}={{10}^{-6}}$, the step size ${{N}_{\text{step}}}=6$, and the initial reservoir size is set to 5. In addition, the grid search method is utilized to find the optimal hyper-parameters, including the scaling factor $\alpha$ in Eq. (\ref{eq22}) and reservoir size $N$. Each experiment is conducted with 50 independent trials under the same conditions. The model performance is evaluated by the means and standard deviations of training and testing NRMSE, as well as training time.

\subsection{Mackey-Glass time-series forecasting}
Mackey-Glass (MG) time-series can be formulated by the following differential equation with time delay, that is,
\begin{equation}
\label{eq70}
\frac{{dy}}{{dn}} = \upsilon y\left( n \right) + \frac{{\alpha y\left( {n - \tau } \right)}}{{1 + y{{\left( {n - \tau } \right)}^{10}}}}.
\end{equation}
As $\tau >16.8$, the above system becomes chaotic, aperiodic, nonconvergent, and divergent. According to  \cite{ref10}, the parameters in Eq. (\ref{eq70}) are set to $\upsilon \text{=}-0.1$, $\alpha \text{=}0.2$, and $\tau =17$. The initial values $\left\{ {y\left( 0 \right), \ldots ,y\left( \tau  \right)} \right\}$ are generated from $\left[ {0.1,1.3} \right]$. $\left\{ {y\left( n \right),y\left( {n - 6} \right),y\left( {n - 12} \right),y\left( {n - 18} \right)} \right\}$, as the model inputs, are used to predict $y\left( {n + 6} \right)$. The second-order Runge-Kutta method is employed to generate 1177 sequence points. In our simulation, samples from time instant 1 to 500 are used to train the network, from 501 to 800 for validation, and the remaining samples for testing. The first 20 samples of each set are washed out. In particular, considering the order uncertainty, we assume that certain orders are unknown and design two experimental setups. In the MG1 task, we select ${{\mathbf{u}}}\left( n \right)={{\left[ y\left( n-6  \right),y\left( n-12  \right),y\left( n-18  \right) \right]}}$ to predict $y\left( {n + 6} \right)$. In the MG2 task, the input is set as ${{\mathbf{u}}}\left( n \right)={{\left[ y\left( n-12  \right),y\left( n-18  \right) \right]}^{\top }}$. These settings are purposely designed to see the power of RSCN with unfilling input variables.
\begin{figure}[ht] 
	\centering
        \includegraphics[width=3.5in]{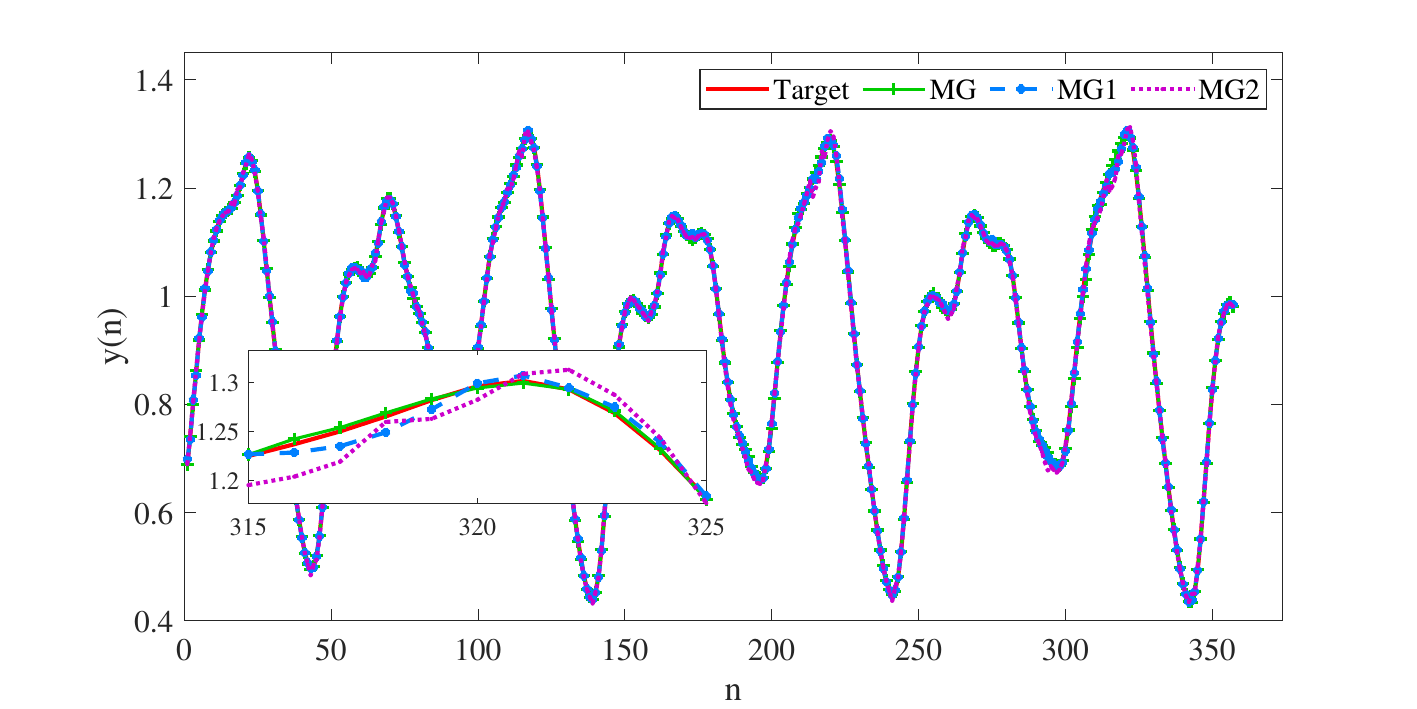}
	\caption{Prediction fitting curves of the RSCN for MG tasks.}
	\label{fig2}
    \vspace{-0.2cm}
\end{figure}

Fig.~\ref{fig2} illustrates the prediction results of RSCN on three MG tasks, where RSCN exhibits superior performance on the MG task compared to MG1 and MG2, indicating that each input variable contributes to the model output. To comprehensively compare the performance of different models in MG time-series prediction tasks, we summarize the results in Table~\ref{tb1}. It can be seen that the NRMSE of each model (excluding LSTM) for MG, MG1, and MG2 increases progressively, suggesting a significant influence of order uncertainty on model performance. The training and testing NRMSE of RSCN are notably lower than other methods. Specifically, compared with the original ESN, the training and testing NRMSE of RSCN are significantly reduced by more than 50\% while using fewer reservoir nodes. These results validate the effectiveness of the proposed RSCNs in constructing a concise reservoir with strong learning capabilities for MG time-series prediction. It is worth noting that LSTM exhibits the longest training time, highlighting the potential of randomized learning algorithms in developing efficient models with reduced computational costs. The training time of RSCN is extended as it involves identifying nodes that meet inequality constraints during incremental construction, resulting in higher time consumption.
\begin{table}[htbp]
\footnotesize
\vspace{-0.5cm}
\caption{Performance comparison of different models on MG tasks.} \label{tb1}
\centering
\setlength\tabcolsep{2pt}
\begin{tabular}{cccccc}
\hline
Datasets             & Models & $N$ & Training time            & Training NRMSE           & \multicolumn{1}{c}{Testing NRMSE} \\ \hline
\multirow{6}{*}{MG}  & LSTM   & 6                  & 6.2316±1.4418          & 0.0327±0.0048          & 0.0389±0.0037                    \\
                     & ESN    & 98                 & 0.1438±0.0378          & 0.0115±0.0021          & 0.0260±0.0083                    \\
                     & SCR    & 79                 & \textbf{0.1308±0.0628} & 0.0070±0.0004          & 0.0153±0.0076                    \\
                     & PESN   & 83                 & 0.5809±0.1857          & 0.0058±0.0007          & 0.0181±0.0102                    \\
                     & LIESN  & 102                & 0.4137±0.1028          & 0.0051±0.0006          & 0.0137±0.0051                    \\
                     & RSCN   & 67                 & 0.9578±0.2071          & \textbf{0.0034±0.0003} & \textbf{0.0121±0.0057}           \\ \hline
\multirow{6}{*}{MG1} & LSTM   & 7                  & 6.1422±1.1742          & 0.0311±0.0080          & 0.0375±0.0077                    \\
                     & ESN    & 124                & 0.1698±0.0508          & 0.0157±0.0067          & 0.0398±0.0113                    \\
                     & SCR    & 103                & \textbf{0.1305±0.0829} & 0.0090±0.0007          & 0.0355±0.0092                    \\
                     & PESN   & 109                & 0.4986±0.1093          & 0.0088±0.0012          & 0.0303±0.0052                    \\
                     & LIESN  & 121                & 0.6699±0.2023          & 0.0078±0.0003          & 0.0255±0.0083                    \\
                     & RSCN   & 79                 & 0.8892±0.5724          & \textbf{0.0051±0.0003} & \textbf{0.0147±0.0035}           \\ \hline
\multirow{6}{*}{MG2} & LSTM   & 5                  & 6.0068±1.1065          & 0.0358±0.0045         & 0.0889±0.0250                    \\
                     & ESN    & 135                & 0.1533±0.0736          & 0.0309±0.0192          & 0.0830±0.0112                    \\
                     & SCR    & 111                & \textbf{0.1268±0.0432} & 0.0213±0.0098          & 0.0766±0.0028                    \\
                     & PESN   & 113                & 0.6302±0.1429          & 0.0249±0.0093          & 0.0822±0.0063                    \\
                     & LIESN  & 126                & 0.4597±0.0837          & 0.0160±0.0063          & 0.0652±0.0029                    \\
                     & RSCN   & 105                & 1.1988±0.2829          & \textbf{0.0071±0.0059} & \textbf{0.0343±0.0028}           \\ \hline
\end{tabular}
\end{table}

\begin{figure} 
	\centering
	\subfloat[MG1]{\includegraphics[width=4.2cm]{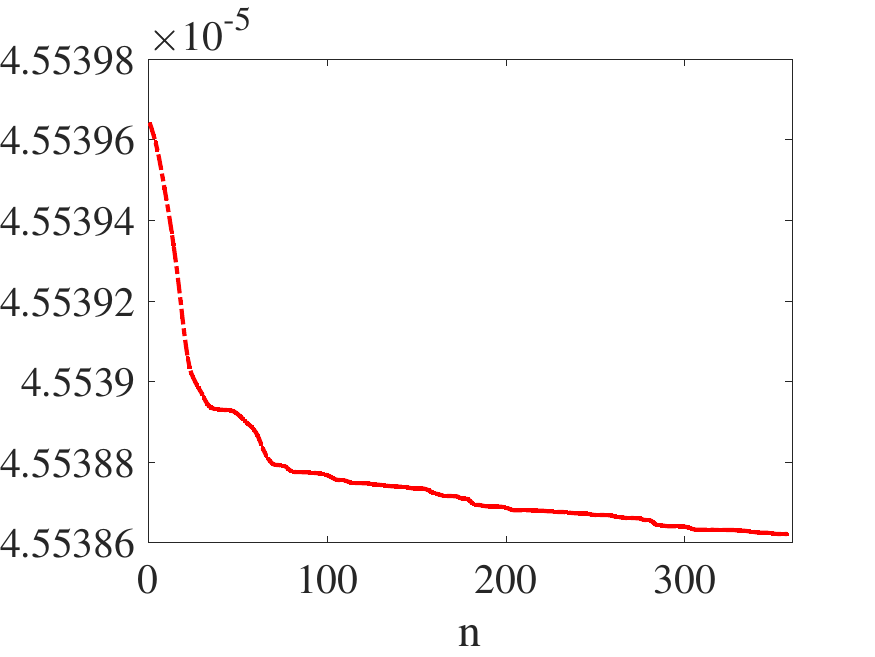}}
        \subfloat[MG2]{\includegraphics[width=4.2cm]{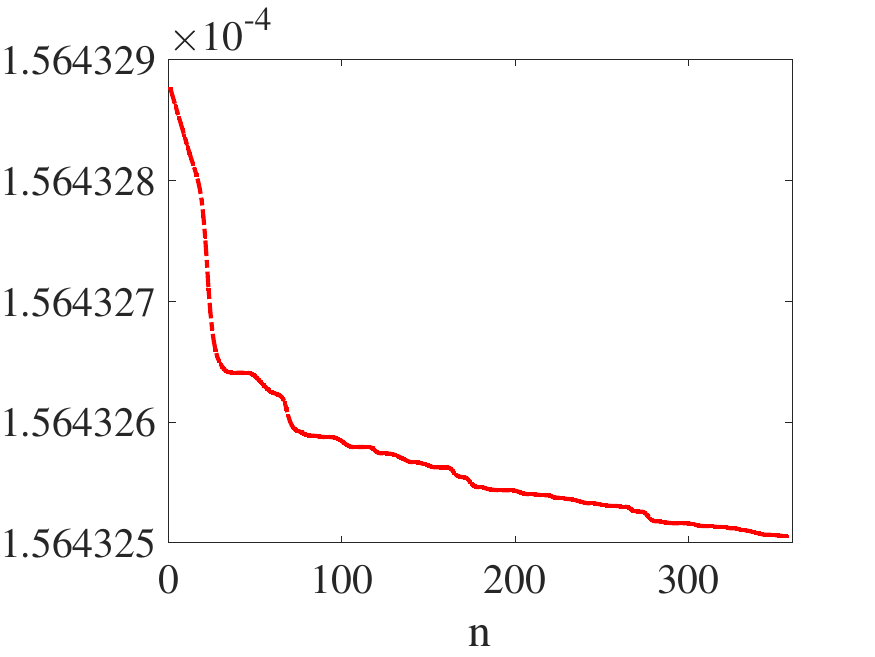}}
	\caption{Errors between the output weights updated by the projection algorithm and trained offline on the MG1 and MG2 tasks.}
	\label{fig3}\vspace{-0.5cm}
\end{figure}

Fig.~\ref{fig3} depicts the errors between the output weights updated by the projection algorithm and the weights trained offline for the MG1 and MG2 tasks. The gradual reduction in error demonstrates the convergence of the output weights.

\subsection{Nonlinear system identification}
In this simulation, the nonlinear system identification problem is considered. The dynamic nonlinear plant can be described as follows
\begin{equation} \label{eq71}
\begin{array}{l}
y\left( {n + 1} \right) = 0.72y\left( n \right) + 0.025y\left( {n - 1} \right)u\left( n \right) + \\
{\kern 1pt} {\kern 1pt} {\kern 1pt} {\kern 1pt} {\kern 1pt} {\kern 1pt} {\kern 1pt} {\kern 1pt} {\kern 1pt} {\kern 1pt} {\kern 1pt} {\kern 1pt} {\kern 1pt} {\kern 1pt} {\kern 1pt} {\kern 1pt} {\kern 1pt} {\kern 1pt} {\kern 1pt} {\kern 1pt} {\kern 1pt} {\kern 1pt} {\kern 1pt} {\kern 1pt} {\kern 1pt} {\kern 1pt} {\kern 1pt} {\kern 1pt} {\kern 1pt} {\kern 1pt} {\kern 1pt} {\kern 1pt} {\kern 1pt} {\kern 1pt} {\kern 1pt} {\kern 1pt} {\kern 1pt} {\kern 1pt} {\kern 1pt} {\kern 1pt} {\kern 1pt} {\kern 1pt} {\kern 1pt} {\kern 1pt} {\kern 1pt} {\kern 1pt} {\kern 1pt} {\kern 1pt} {\kern 1pt} {\kern 1pt} 0.01{u^2}\left( {n - 2} \right) + 0.2u\left( {n - 3} \right).
\end{array}
\end{equation}
In the training phase, $u\left( n \right)$ is generated from the uniform distribution $\left[ { - 1,1} \right]$, and the initial output $y\left( 1 \right)=y\left( 2 \right)=y\left( 3 \right)=0, y\left( 4 \right)=0.1.$ In the testing phase, the input is
\begin{equation}
\label{eq72}
u\left( n \right) = \left\{ {\begin{array}{*{20}{l}}
{\sin \left( {\frac{{\pi n}}{{25}}} \right),{\kern 1pt} {\kern 1pt} {\kern 1pt} {\kern 1pt} {\kern 1pt} {\kern 1pt} {\kern 1pt} {\kern 1pt} {\kern 1pt} {\kern 1pt} {\kern 1pt} {\kern 1pt} {\kern 1pt} {\kern 1pt} {\kern 1pt} {\kern 1pt} {\kern 1pt} {\kern 1pt} {\kern 1pt} {\kern 1pt} {\kern 1pt} {\kern 1pt} {\kern 1pt} {\kern 1pt} {\kern 1pt} {\kern 1pt} {\kern 1pt} {\kern 1pt} {\kern 1pt} {\kern 1pt} {\kern 1pt} {\kern 1pt} {\kern 1pt} {\kern 1pt} {\kern 1pt} {\kern 1pt} {\kern 1pt} {\kern 1pt} {\kern 1pt} {\kern 1pt} {\kern 1pt} {\kern 1pt} {\kern 1pt} {\kern 1pt} {\kern 1pt} {\kern 1pt} {\kern 1pt} {\kern 1pt} {\kern 1pt} {\kern 1pt} {\kern 1pt} {\kern 1pt} {\kern 1pt} {\kern 1pt} {\kern 1pt} {\kern 1pt} {\kern 1pt} {\kern 1pt} {\kern 1pt} {\kern 1pt} {\kern 1pt} {\kern 1pt} {\kern 1pt} {\kern 1pt} {\kern 1pt} {\kern 1pt} {\kern 1pt} {\kern 1pt} {\kern 1pt} {\kern 1pt} {\kern 1pt} {\kern 1pt} {\kern 1pt} {\kern 1pt} {\kern 1pt} {\kern 1pt} {\kern 1pt} {\kern 1pt} {\kern 1pt} {\kern 1pt} {\kern 1pt} {\kern 1pt} {\kern 1pt} {\kern 1pt} {\kern 1pt} {\kern 1pt} {\kern 1pt} {\kern 1pt} {\kern 1pt} {\kern 1pt} {\kern 1pt} {\kern 1pt} {\kern 1pt} {\kern 1pt} {\kern 1pt} {\kern 1pt} {\kern 1pt} {\kern 1pt} {\kern 1pt} {\kern 1pt} {\kern 1pt} {\kern 1pt} 0 < n < 250}\\
{1,{\kern 1pt}  {\kern 1pt} {\kern 1pt} {\kern 1pt} {\kern 1pt} {\kern 1pt} {\kern 1pt} {\kern 1pt} {\kern 1pt} {\kern 1pt} {\kern 1pt} {\kern 1pt} {\kern 1pt} {\kern 1pt} {\kern 1pt} {\kern 1pt} {\kern 1pt} {\kern 1pt} {\kern 1pt} {\kern 1pt} {\kern 1pt} {\kern 1pt} {\kern 1pt} {\kern 1pt} {\kern 1pt} {\kern 1pt} {\kern 1pt} {\kern 1pt} {\kern 1pt} {\kern 1pt} {\kern 1pt} {\kern 1pt} {\kern 1pt} {\kern 1pt} {\kern 1pt} {\kern 1pt} {\kern 1pt} {\kern 1pt} {\kern 1pt} {\kern 1pt} {\kern 1pt} {\kern 1pt} {\kern 1pt} {\kern 1pt} {\kern 1pt} {\kern 1pt} {\kern 1pt} {\kern 1pt} {\kern 1pt} {\kern 1pt} {\kern 1pt} {\kern 1pt} {\kern 1pt} {\kern 1pt} {\kern 1pt} {\kern 1pt}{\kern 1pt} {\kern 1pt} {\kern 1pt} {\kern 1pt} {\kern 1pt} {\kern 1pt} {\kern 1pt} {\kern 1pt} {\kern 1pt} {\kern 1pt} {\kern 1pt} {\kern 1pt} {\kern 1pt} {\kern 1pt} {\kern 1pt} {\kern 1pt} {\kern 1pt} {\kern 1pt} {\kern 1pt} {\kern 1pt} {\kern 1pt} {\kern 1pt} {\kern 1pt} {\kern 1pt} {\kern 1pt} {\kern 1pt} {\kern 1pt} {\kern 1pt} {\kern 1pt} {\kern 1pt} {\kern 1pt} {\kern 1pt} {\kern 1pt} {\kern 1pt} {\kern 1pt} {\kern 1pt} {\kern 1pt} {\kern 1pt} {\kern 1pt} {\kern 1pt} {\kern 1pt} {\kern 1pt} {\kern 1pt} {\kern 1pt} {\kern 1pt} {\kern 1pt} {\kern 1pt} {\kern 1pt} {\kern 1pt} {\kern 1pt} {\kern 1pt} {\kern 1pt} {\kern 1pt} {\kern 1pt} {\kern 1pt} {\kern 1pt} {\kern 1pt} {\kern 1pt} {\kern 1pt} {\kern 1pt} {\kern 1pt} {\kern 1pt} {\kern 1pt} {\kern 1pt} {\kern 1pt} {\kern 1pt} {\kern 1pt} {\kern 1pt} {\kern 1pt} {\kern 1pt} 250 \le n < 500}\\
{ - 1,{\kern 1pt}  {\kern 1pt} {\kern 1pt}  {\kern 1pt} {\kern 1pt} {\kern 1pt} {\kern 1pt} {\kern 1pt} {\kern 1pt} {\kern 1pt}{\kern 1pt} {\kern 1pt} {\kern 1pt} {\kern 1pt} {\kern 1pt} {\kern 1pt} {\kern 1pt} {\kern 1pt} {\kern 1pt} {\kern 1pt} {\kern 1pt} {\kern 1pt} {\kern 1pt} {\kern 1pt} {\kern 1pt} {\kern 1pt} {\kern 1pt} {\kern 1pt} {\kern 1pt} {\kern 1pt} {\kern 1pt} {\kern 1pt} {\kern 1pt} {\kern 1pt} {\kern 1pt} {\kern 1pt} {\kern 1pt} {\kern 1pt} {\kern 1pt} {\kern 1pt} {\kern 1pt} {\kern 1pt} {\kern 1pt} {\kern 1pt} {\kern 1pt} {\kern 1pt} {\kern 1pt} {\kern 1pt} {\kern 1pt} {\kern 1pt} {\kern 1pt} {\kern 1pt} {\kern 1pt} {\kern 1pt} {\kern 1pt} {\kern 1pt} {\kern 1pt} {\kern 1pt} {\kern 1pt} {\kern 1pt} {\kern 1pt} {\kern 1pt} {\kern 1pt} {\kern 1pt} {\kern 1pt} {\kern 1pt} {\kern 1pt} {\kern 1pt} {\kern 1pt} {\kern 1pt} {\kern 1pt} {\kern 1pt} {\kern 1pt} {\kern 1pt} {\kern 1pt} {\kern 1pt} {\kern 1pt} {\kern 1pt} {\kern 1pt} {\kern 1pt} {\kern 1pt} {\kern 1pt} {\kern 1pt} {\kern 1pt} {\kern 1pt} {\kern 1pt} {\kern 1pt} {\kern 1pt} {\kern 1pt} {\kern 1pt} {\kern 1pt} {\kern 1pt} {\kern 1pt} {\kern 1pt} {\kern 1pt} {\kern 1pt} {\kern 1pt} {\kern 1pt} {\kern 1pt} {\kern 1pt} {\kern 1pt} {\kern 1pt} {\kern 1pt} {\kern 1pt} {\kern 1pt} {\kern 1pt} {\kern 1pt} {\kern 1pt} {\kern 1pt} {\kern 1pt} {\kern 1pt} {\kern 1pt} {\kern 1pt} {\kern 1pt} {\kern 1pt} {\kern 1pt} {\kern 1pt} {\kern 1pt} 500 \le n < 750}\\
0.6\cos \left( {\frac{{\pi n}}{{10}}} \right) + 0.1\cos \left( {\frac{{\pi n}}{{32}}} \right) + \\
0.3\sin \left( {\frac{{\pi n}}{{25}}} \right),{\kern 1pt} {\kern 1pt} {\kern 1pt} {\kern 1pt} {\kern 1pt} {\kern 1pt} {\kern 1pt} {\kern 1pt} {\kern 1pt} {\kern 1pt} {\kern 1pt} {\kern 1pt} {\kern 1pt} {\kern 1pt} {\kern 1pt} {\kern 1pt} {\kern 1pt} {\kern 1pt}  {\kern 1pt} {\kern 1pt} {\kern 1pt} {\kern 1pt} {\kern 1pt} {\kern 1pt}  {\kern 1pt} {\kern 1pt} {\kern 1pt}{\kern 1pt} {\kern 1pt} {\kern 1pt} {\kern 1pt} {\kern 1pt} {\kern 1pt} {\kern 1pt} {\kern 1pt} {\kern 1pt} {\kern 1pt} {\kern 1pt} {\kern 1pt} {\kern 1pt} {\kern 1pt} {\kern 1pt} {\kern 1pt} {\kern 1pt} {\kern 1pt} {\kern 1pt} {\kern 1pt} {\kern 1pt} {\kern 1pt} {\kern 1pt} {\kern 1pt} {\kern 1pt} {\kern 1pt} {\kern 1pt} {\kern 1pt} {\kern 1pt} {\kern 1pt} {\kern 1pt} {\kern 1pt} {\kern 1pt} {\kern 1pt} {\kern 1pt} {\kern 1pt} {\kern 1pt} {\kern 1pt} {\kern 1pt} {\kern 1pt} {\kern 1pt} {\kern 1pt} {\kern 1pt} {\kern 1pt} {\kern 1pt} {\kern 1pt} {\kern 1pt} {\kern 1pt} {\kern 1pt} {\kern 1pt} 750 \le n \le 1000.
\end{array}
} \right.
\end{equation}
The output of the nonlinear system is predicted by ${{\left[ y\left( n \right),u\left( n \right) \right]}^{\top }}$. The dataset consists of 4000 samples, divided into 2000 for training, 1000 for validation, and 1000 for testing. The initial 100 samples of each set are washed out.\vspace{-0.5cm}
\begin{figure}[ht] 
	\centering
        \includegraphics[width=3.5in]{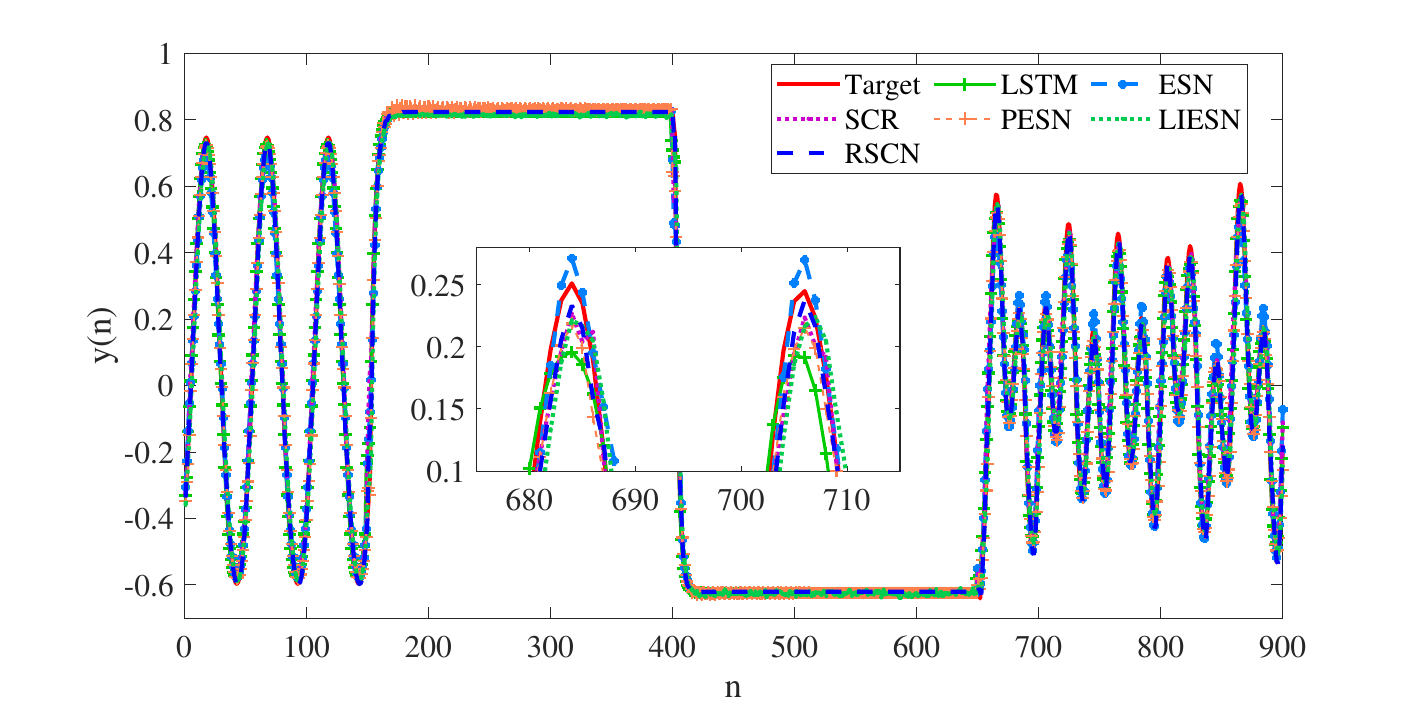}
	\caption{The prediction curves of each model for the nonlinear system identification task.}
	\label{fig601}
\end{figure}

Fig.~\ref{fig601} depicts the prediction results of each model for the nonlinear system identification task. The model outputs of RSCNs show a closer match with the desired outputs in comparison to other models. These findings demonstrate that RSCNs can quickly adapt to changes in nonlinear dynamic systems, thus confirming the effectiveness of the proposed approach.
\begin{figure}[ht] 
	\centering
	\subfloat{\includegraphics[width=8cm]{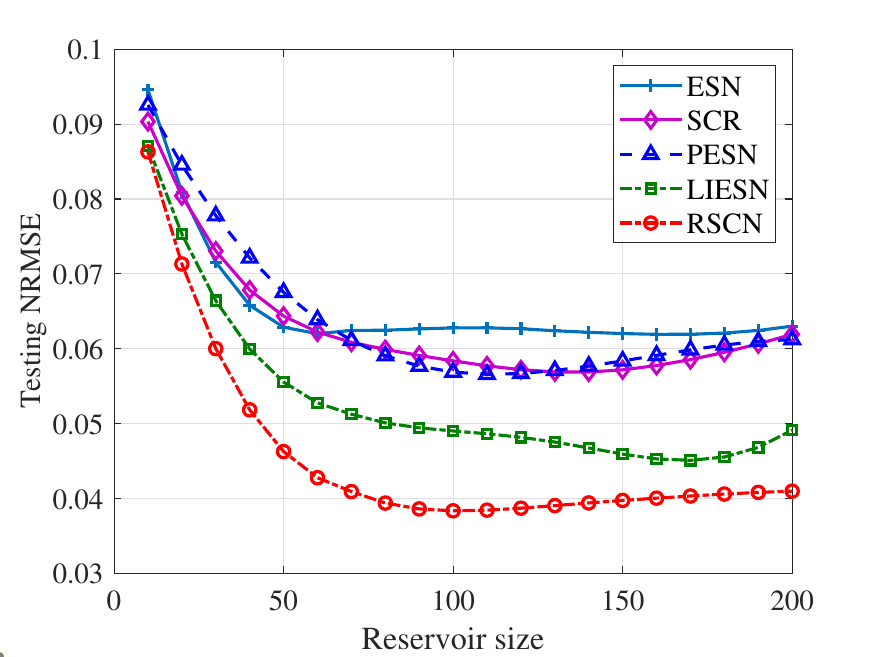}}
	\caption{Performance comparison of various models with different reservoir sizes on the nonlinear system identification task.}
	\label{fig701}
\end{figure}

To investigate the impact of reservoir size on the model performance, we evaluate the testing performance with different reservoir sizes. Fig.~\ref{fig701} illustrates how testing NRMSE varies across reservoir sizes for each method. It is clear that an excessive number of reservoir nodes can lead to overfitting in the ESN, SCR, PESN, LIESN, and RSCN when reservoir sizes are 170, 140, 120, 170, and 100, respectively. Therefore, constructing an effective and concise reservoir topology is crucial for improving tracking performance. RSCN stands out for its ability to generate a suitable network structure by randomly assigning weights and biases through a supervisory mechanism, which also guarantees its universal approximation property.

Table~\ref{tb3} presents a comprehensive performance analysis for the nonlinear system identification task. RSCN consistently outperforms LSTM, ESN, SCR, PESN, and LIESN in both training and testing NRMSE. Specifically, the training and testing NRMSE obtained by RSCN accounts for 86.68\% and 63.07\% of those obtained by ESN. The superiority of RSCN is further highlighted by its smaller reservoir size compared to other models, which accounts for 64.97\% of those obtained by ESN. This suggests that RSCN can achieve comparable prediction results with fewer nodes, resulting in a more compact reservoir. Moreover, compared to other models, the proposed RSCN exhibits a superior ability to identify complex patterns and dependencies within the system, confirming its effectiveness in nonlinear system identification.
\begin{table}[htbp]
\vspace{-0.3cm}
\centering
\caption{Performance comparison of different models on the nonlinear system identification task.} \label{tb3}
\begin{tabular}{ccccc}
\hline
Models & $N$ & Training time    & Training NRMSE  & Testing NRMSE   \\ \hline
LSTM   & 19                 & 19.2813±6.4821 & 0.0096±0.0173 & 0.0555±0.0274 \\
ESN    & 157                &\textbf{0.7633±0.0103}  & 0.0091±0.0025 & 0.0627±0.0032 \\
SCR    & 136                & 0.8838±0.0026  & 0.0084±0.0006 & 0.0588±0.0063 \\
PESN   & 118                & 6.0845±0.9190  & 0.0082±0.0008 & 0.0584±0.0049 \\
LIESN  & 153                & 1.2621±0.0377  & 0.0081±0.0005 & 0.0469±0.0070 \\
RSCN   & 102                & 12.0336±0.8535 & \textbf{0.0079±0.0005} & \textbf{0.0395±0.0009} \\ \hline
\end{tabular}
\end{table}

\begin{figure}[ht] 
	\centering
	\subfloat{\includegraphics[width=5.2cm]{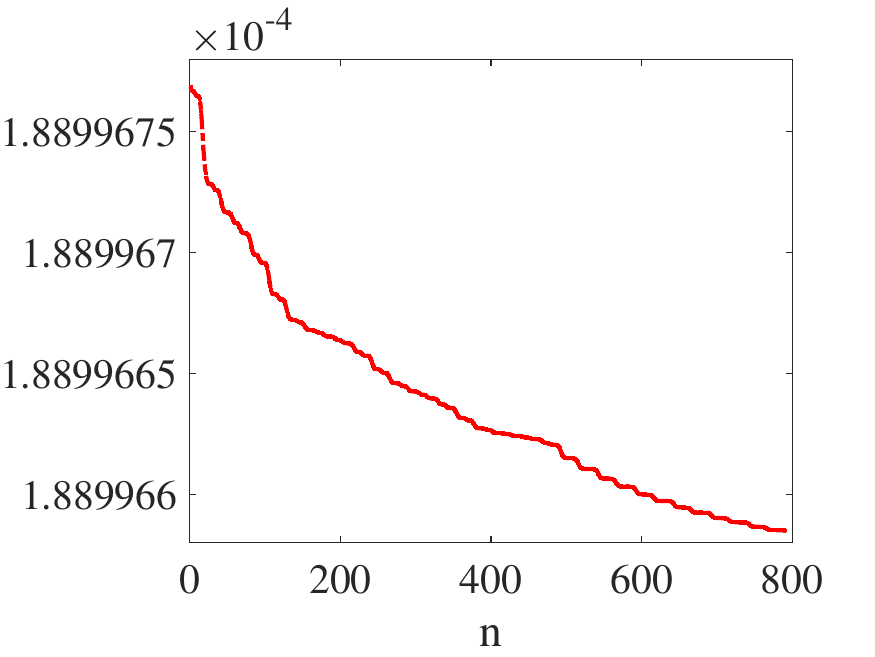}}
	\caption{Errors between the output weights updated by the projection algorithm and the weights trained offline on the nonlinear system identification task.}
	\label{fig801}
    \vspace{-0.5cm}
\end{figure}

The errors between the output weights updated by the projection algorithm and the weights trained offline for the nonlinear system identification task are shown in Fig.~\ref{fig801}. It is evident that the output weights are converging, indicating that the model is approaching a stable state. This stability is crucial for the precise identification of nonlinear systems, ensuring the model’s reliability and consistency when encountering new data or scenarios.

\subsection{Soft sensing of butane in dehumanizer column}
\begin{figure}[ht]
	\begin{center}
		\includegraphics[width=6cm]{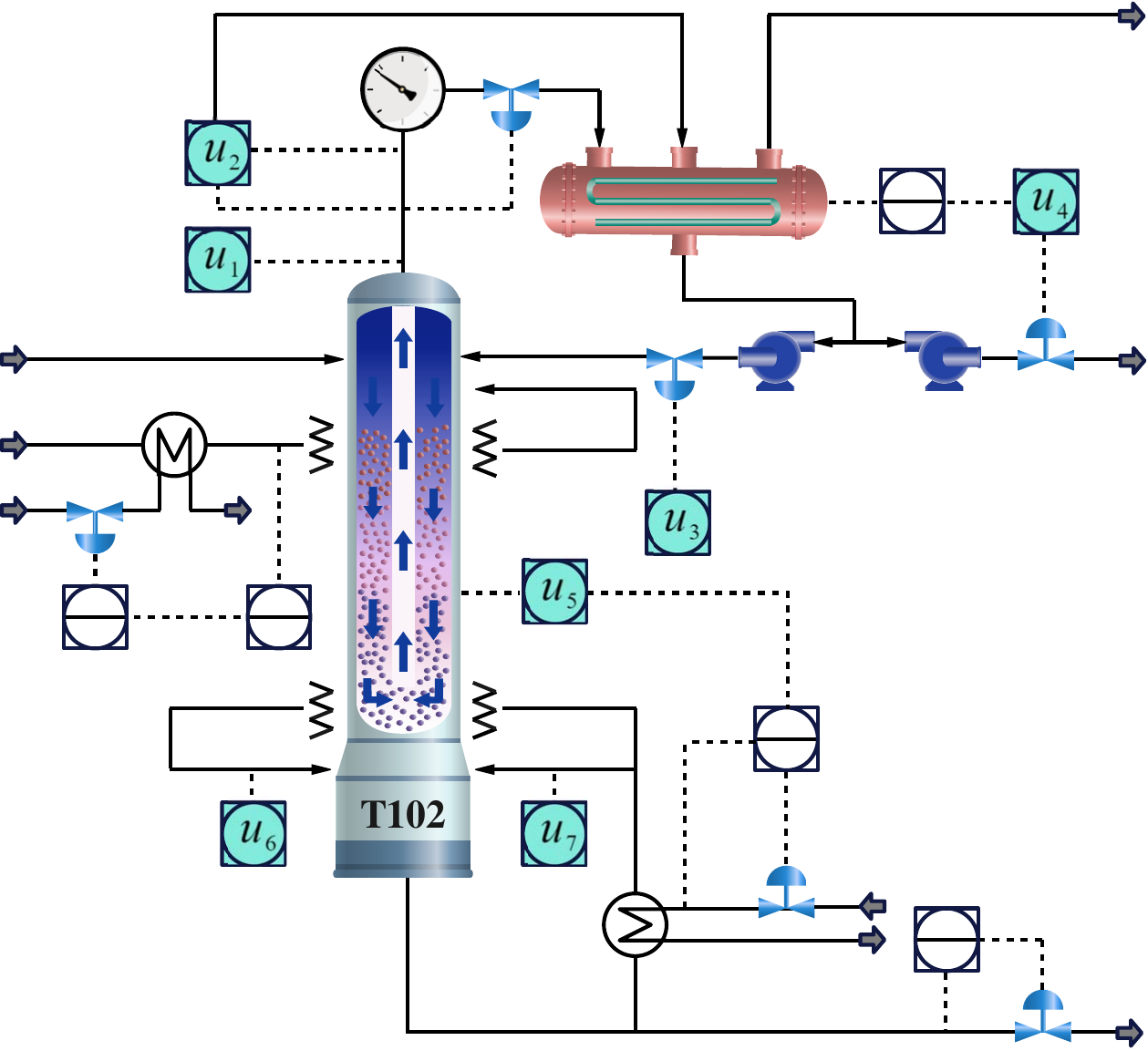}
		\caption{Flowchart of the debutanizer column.}
		\label{fig4}
	\end{center}
\end{figure}

The debutanizer column plays a crucial role in the desulfurization and naphtha separation unit within petroleum refining production. The flowchart is shown in Fig.~\ref{fig4}. There are seven auxiliary variables in the refining process, which are described as tower top temperature $u_{1}$, tower top pressure $u_{2}$, tower top reflux flow $u_{3}$, tower top product outflow $u_{4}$, 6-th tray temperature $u_{5}$, tower bottom temperature $u_{6}$, and tower bottom pressure $u_{7}$. One dominant variable $y$ is the butane concentration at the bottom of the tower, which cannot be directly detected and needs to meet the quality control requirements of minimizing content. This variable is typically obtained through online instrument analysis. The data are generated from a real-time sampling process, with 2394 samples taken at 6-minute intervals. The previous study in \cite{ref36} has highlighted the delay problem in the debutanizer column, which could be resolved by introducing several well-designed new variables into the original seven variables, that is,
\begin{equation} \label{eq71}
\begin{array}{l}
y\left( n \right) = f\left( {{u_1}\left( n \right),} \right.{u_2}\left( n \right),{u_3}\left( n \right),{u_4}\left( n \right),{u_5}\left( n \right),{u_5}\left( {n - 1} \right),\\
{\kern 1pt} {\kern 1pt} {\kern 1pt} {\kern 1pt} {\kern 1pt} {\kern 1pt} {\kern 1pt} {\kern 1pt} {\kern 1pt} {\kern 1pt} {\kern 1pt} {\kern 1pt} {\kern 1pt} {\kern 1pt} {\kern 1pt} {\kern 1pt} {\kern 1pt} {\kern 1pt} {\kern 1pt} {\kern 1pt} {\kern 1pt} {\kern 1pt} {\kern 1pt} {\kern 1pt} {\kern 1pt} {\kern 1pt} {\kern 1pt} {\kern 1pt} {\kern 1pt} {\kern 1pt} {\kern 1pt} {\kern 1pt} {\kern 1pt} {\kern 1pt} {\kern 1pt} {\kern 1pt} {\kern 1pt} {\kern 1pt} {\kern 1pt} {\kern 1pt} {\kern 1pt} {\kern 1pt} {\kern 1pt} {\kern 1pt} {\kern 1pt} {\kern 1pt} {\kern 1pt} {\kern 1pt} {u_5}\left( {n - 2} \right),{u_5}\left( {n - 3} \right),\left( {{u_1}\left( n \right) + {u_2}\left( n \right)} \right)/2,\\
{\kern 1pt} {\kern 1pt} {\kern 1pt} {\kern 1pt} {\kern 1pt} {\kern 1pt} {\kern 1pt} {\kern 1pt} {\kern 1pt} {\kern 1pt} {\kern 1pt} {\kern 1pt} {\kern 1pt} {\kern 1pt} {\kern 1pt} {\kern 1pt} {\kern 1pt} {\kern 1pt} {\kern 1pt} {\kern 1pt} {\kern 1pt} {\kern 1pt} {\kern 1pt} {\kern 1pt} {\kern 1pt} {\kern 1pt} {\kern 1pt} {\kern 1pt} {\kern 1pt} {\kern 1pt} {\kern 1pt} {\kern 1pt} {\kern 1pt} {\kern 1pt} {\kern 1pt} {\kern 1pt} {\kern 1pt} {\kern 1pt} {\kern 1pt} {\kern 1pt} {\kern 1pt} {\kern 1pt} {\kern 1pt} {\kern 1pt} {\kern 1pt} {\kern 1pt} {\kern 1pt} \left. {y\left( {n - 1} \right),y\left( {n - 2} \right),y\left( {n - 3} \right),y\left( {n - 4} \right)} \right).
\end{array}
\end{equation}Specifically, considering the order uncertainty, $\left[ {{u}_{1}}\left( n \right), \right.$ $\left. {{u}_{2}}\left( n \right),{{u}_{3}}\left( n \right),{{u}_{4}}\left( n \right),{{u}_{5}}\left( n \right),y\left( n-1 \right) \right]$ is used to predict the output $y\left( n \right)$. The first 1500 samples are used for training the model, while the next 894 samples are chosen for testing. Gaussian white noise is added to the testing set to generate the validation set. The first 100 samples are washed out.

\begin{figure}[ht] 
	\centering
        \includegraphics[width=3.5in]{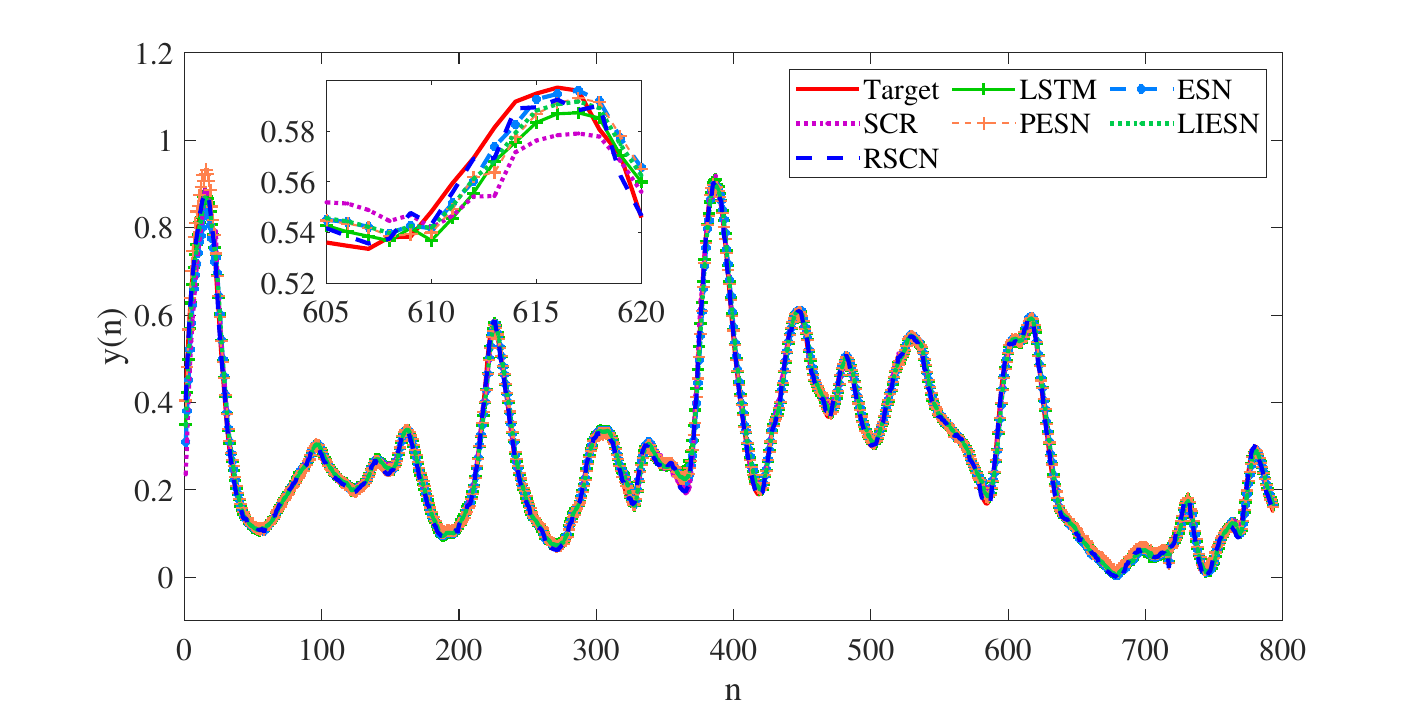}
	\caption{The prediction curves of each model for the dehumanizer column process.}
	\label{fig5}
\end{figure}

Fig.~\ref{fig5} presents the prediction curves of each model for the dehumanizer column process. The RSCN can effectively follow the target output with a higher fitting degree. These results confirm the efficacy of the proposed method in handling real industrial processes with unknown dynamical orders.

\vspace{-0.5cm}

\subsection{Short-term power load forecasting}
\begin{figure}[htbp]
	\begin{center}
		\includegraphics[width=8cm]{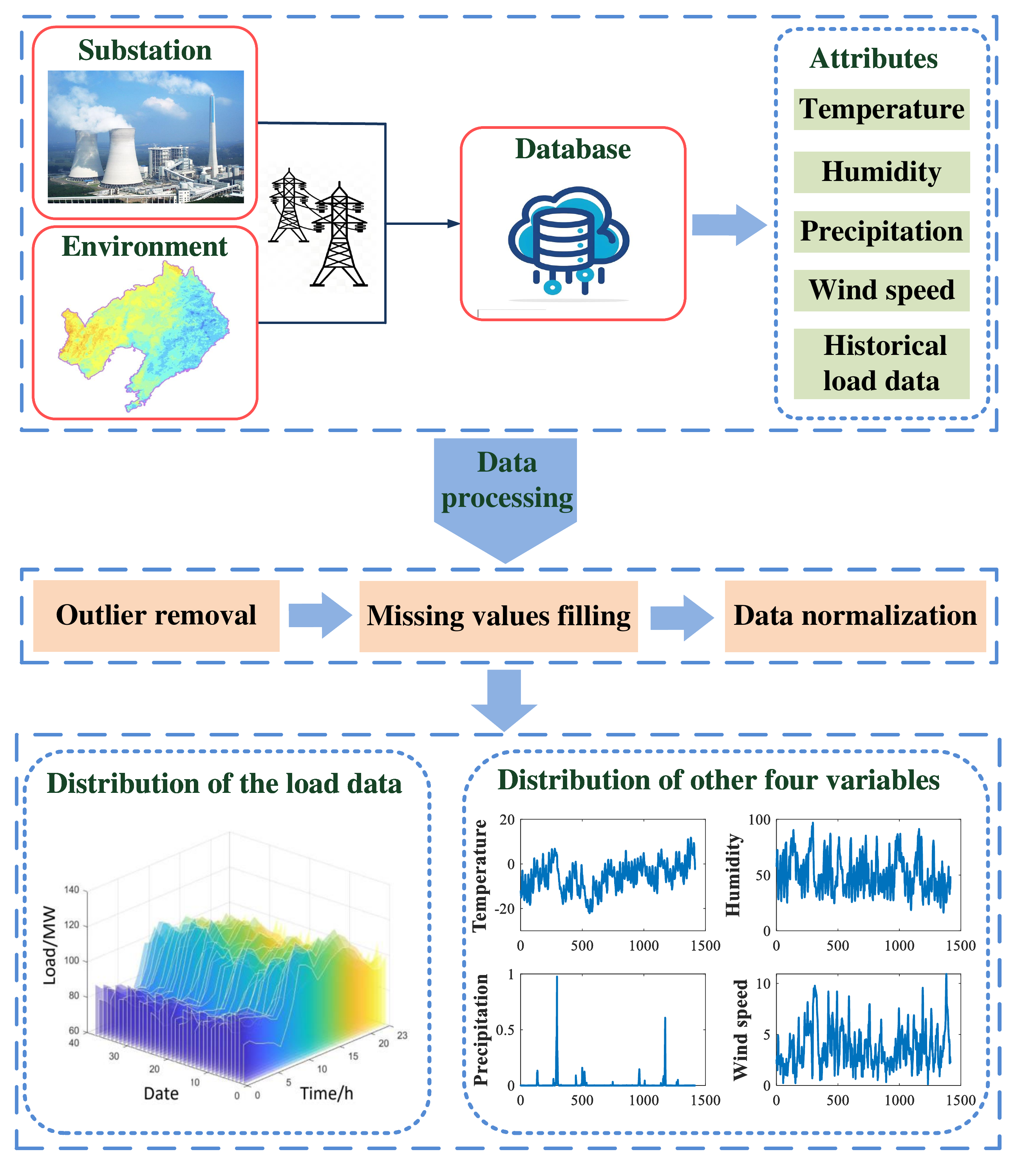}
		\caption{The flowchart of the data collection and processing for the short-term power load forecasting.}
		\label{fig6}
	\end{center}
\end{figure}
Short-term power load forecasting is important for maintaining the safe and stable operation of the power system, enhancing energy efficiency, and minimizing operating costs. The load data used in this work originates from a 500kV substation located in Liaoning, China. Hourly load data was collected from January to February 2023, providing a dataset spanning 59 days. There are four environmental variables including temperature ${{u}_{1}}$, humidity ${{u}_{2}}$, precipitation ${{u}_{3}}$, and wind speed ${{u}_{4}}$ were generated to predict power load $y$. The flowchart of the data collection and processing for the short-term power load forecasting is shown in Fig.~\ref{fig6}. This dataset consists of 1415 samples, with the first 1000 samples allocated for training and the remaining 415 samples for testing. Gaussian noise is introduced to the testing set to create the validation set. Considering the order uncertainty $\left[ {{u}_{1}}\left( n \right), \right.$$\left. {{u}_{2}}\left( n \right),{{u}_{3}}\left( n \right),{{u}_{4}}\left( n \right),y\left( n-1 \right) \right]$ is utilized to predict $y\left( n \right)$ in the experiment. The first 30 samples of each set are washed out.

\begin{figure}[ht] 
	\centering
        \includegraphics[width=3.5in]{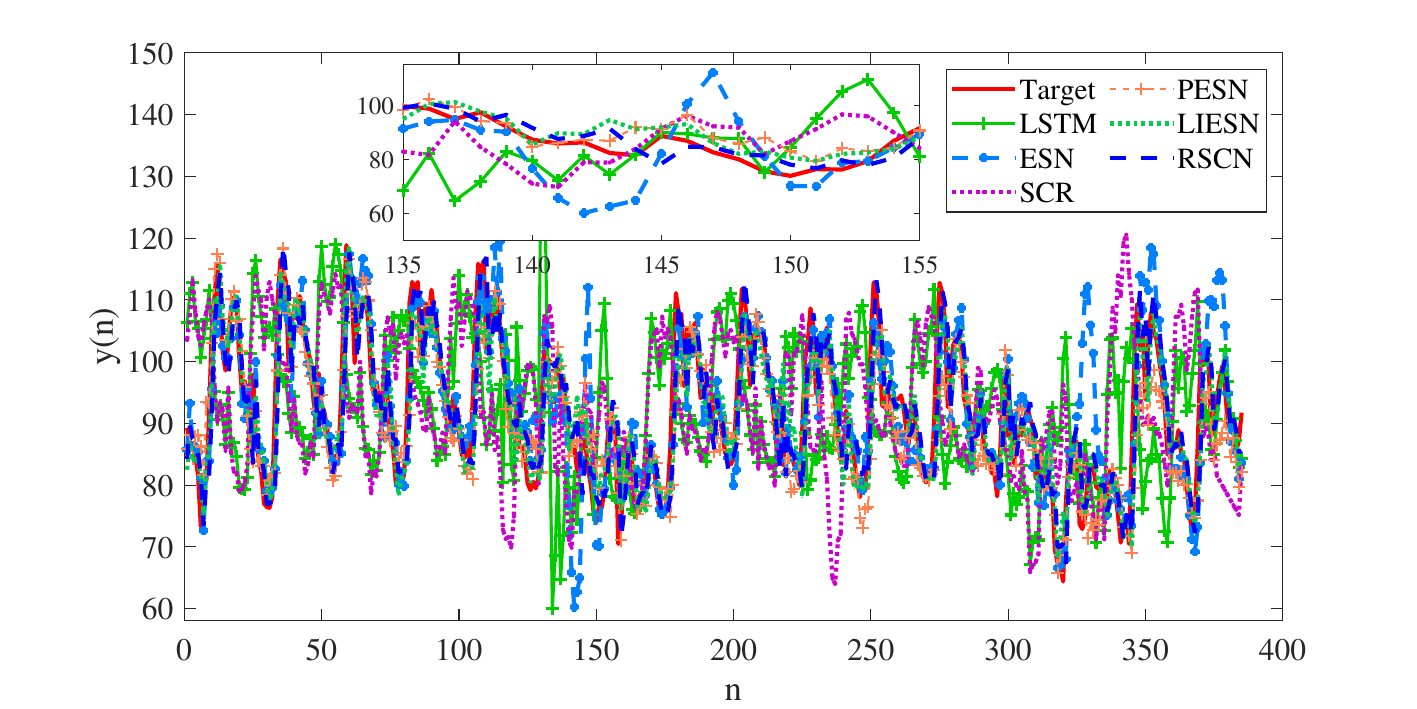}
	\caption{The prediction curves of each model for the short-term power load forecasting.}
	\label{fig7}
\end{figure}

 Fig.~\ref{fig7} depicts the prediction results of different models for the short-term power load forecasting, indicating that the output of RSCN aligns more closely with the desired output. In contrast, other models exhibit significantly larger prediction errors, particularly during periods of substantial data fluctuations or complex changes. These findings suggest that the RSCN model demonstrates superior adaptability and stability in processing nonlinear and dynamically changing temporal data.
 
\vspace{-0.5cm}

\begin{table}[htbp]
\footnotesize
\caption{Performance comparison of different models on the two industry cases.} \label{tb4}
\centering
\setlength\tabcolsep{2pt} 
\begin{tabular}{cccccc}
\hline
Datasets                & Models & $N$ & Training time                                                                                        & Training NRMSE           & Testing NRMSE            \\ \hline
\multirow{6}{*}{Case 1} & LSTM   & 6                  & 23.0278±5.5847                                                                                     & 0.0306±0.0016          & 0.0694±0.0184          \\
                        & ESN    & 213                & 0.4081±0.1203 & 0.0408±0.0013          & 0.0843±0.0114          \\
                        & SCR    & 196                & \textbf{0.2542±0.0314}                                                                             & 0.0312±0.0009          & 0.0778±0.0132          \\
                        & PESN   & 178                & 1.5453±0.1408                                                                                      & 0.0320±0.0011          & 0.0757±0.0116          \\
                        & LIESN  & 153                & 0.9620±0.1079                                                                                      & 0.0281±0.0012          & 0.0715±0.0102          \\
                        & RSCN   & 87                 & 2.2373±0.2012                                                                                      & \textbf{0.0273±0.0010} & \textbf{0.0679±0.0041} \\ \hline
\multirow{6}{*}{Case 2} & LSTM   & 5                  & 18.3910±3.2662                                                                                     & 0.3825±0.0779          & 0.7571±0.0548         \\
                        & ESN    & 103                & 0.1188±0.0154 & 0.2433±0.0218          & 0.6452±0.0717          \\
                        & SCR    & 82                 & \textbf{0.0887±0.0092}                                                                             & 0.2271±0.0147          & 0.6235±0.0432          \\
                        & PESN   & 156                & 0.8273±0.0312                                                                                      & 0.2102±0.0200          & 0.6025±0.0452          \\
                        & LIESN  & 102                & 0.6127±0.0219                                                                                      & 0.2006±0.0211          & 0.5754±0.0382          \\
                        & RSCN   & 65                 & 0.6825±0.0314                                                                                      & \textbf{0.1925±0.0204} & \textbf{0.5107±0.0368} \\ \hline
\end{tabular}
\end{table}

\begin{figure}[ht] 
	\centering
	\subfloat[Case 1]{\includegraphics[width=8cm]{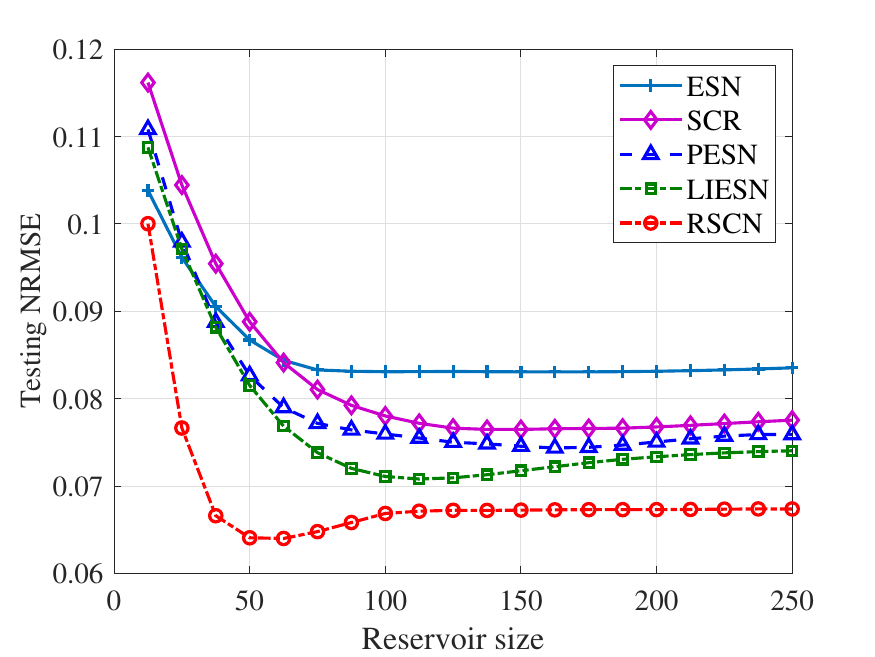}}\\
	\subfloat[Case 2]{\includegraphics[width=8cm]{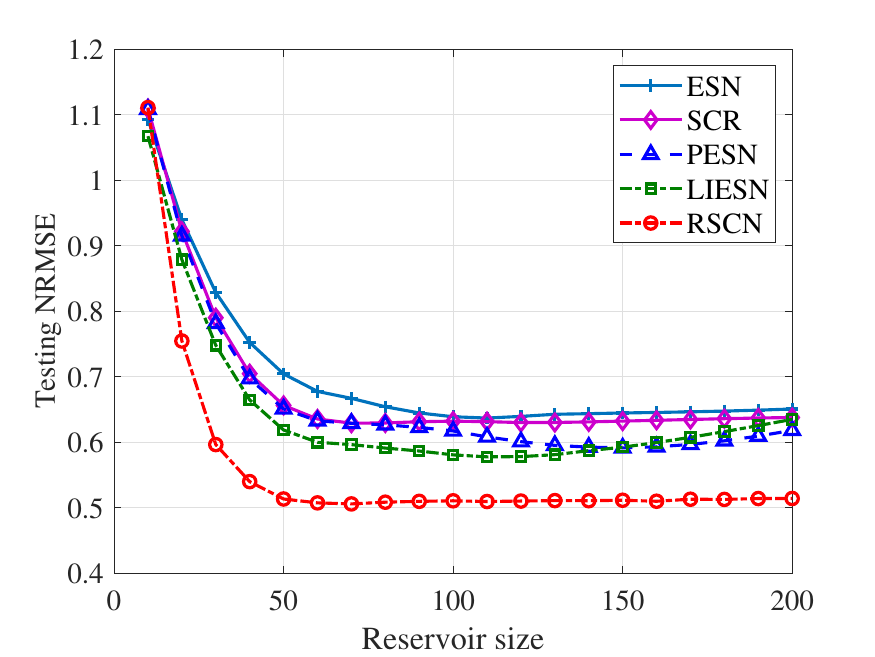}}
	\caption{Performance comparison of various models with different reservoir sizes on the two industry cases.}
	\label{fig8}
\end{figure}

Fig.~\ref{fig8} illustrates the influence of reservoir size on the testing performance for various models. It can be seen that the testing NRMSE of RSCN consistently outperforms that of the other models, particularly when the reservoir pool is small. In such scenarios, the RSCN model maintains a lower NRMSE, while other models' performance experiences significant degradation. This finding demonstrates that RSCN can provide reliable prediction results even in resource-constrained environments. Furthermore, overfitting is observed, with the optimal reservoir sizes for ESN, SCR, PESN, LIESN, and RSCN being 210, 160, 180, 110, and 60 in case 1, and 110, 80, 150, 110, and 70 in case 2. Each reservoir node of RSCN is generated under a supervisory mechanism, allowing it to approach the target output infinitely. This unique learning mechanism ensures that the network can effectively capture key process information in complex dynamics, which is crucial for the analysis and modelling of industrial temporal data.

\begin{figure}[ht] 
	\centering
	\subfloat[Case 1]{\includegraphics[width=4.2cm]{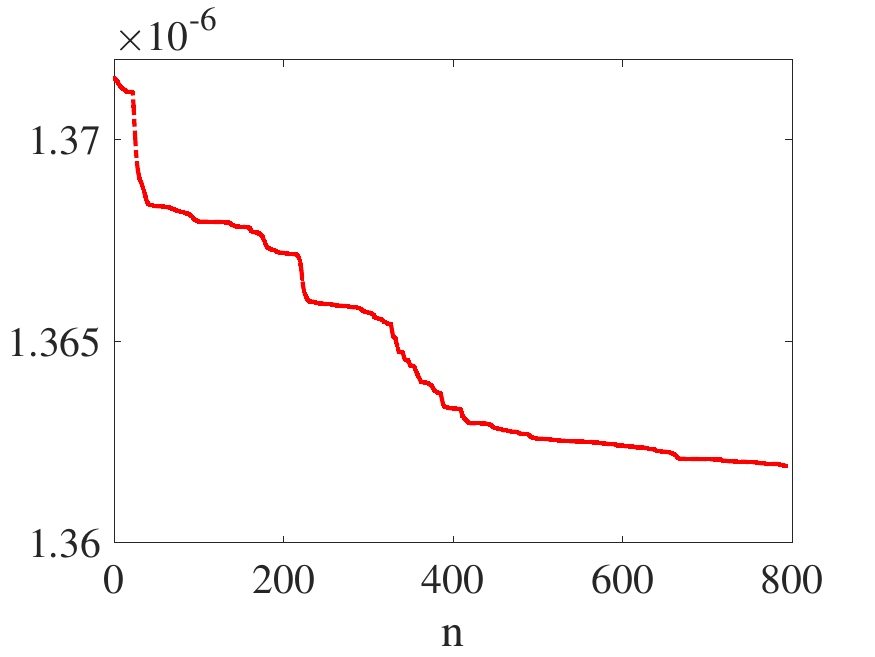}}
	\subfloat[Case 2]{\includegraphics[width=4.2cm]{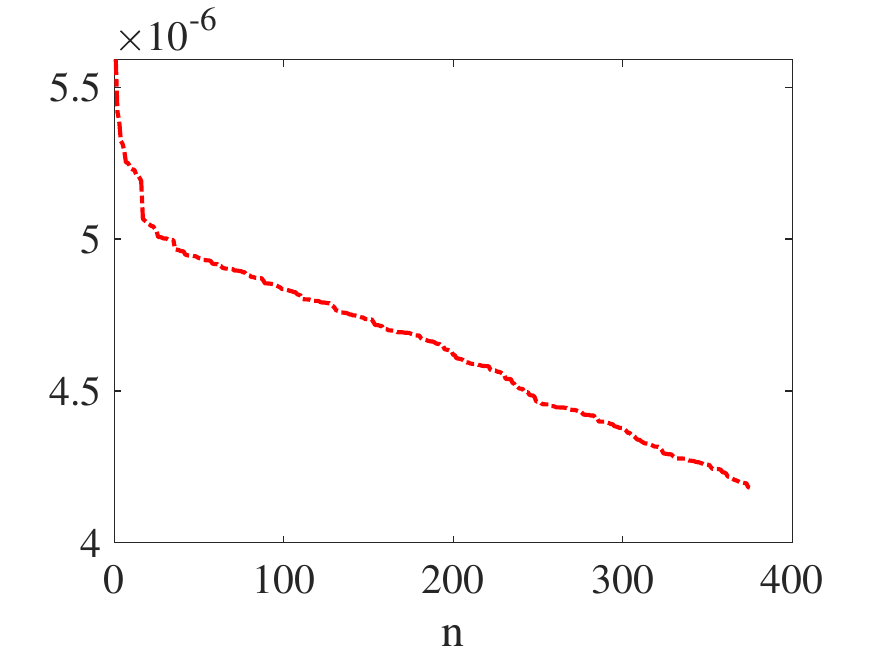}}
	\caption{Errors between the output weights updated by the projection algorithm and trained offline on the two industry cases.}
	\label{fig9}
\end{figure}

To comprehensively compare the modelling performance of the proposed RSCN with other classical models, we present the experimental results in Table~\ref{tb4}. It can be seen that the training and testing NRMSE of RSCN are lower than other models. In particular, RSCN features a more compact reservoir topology, which can improve computing efficiency and reduce practical resource costs. Fig.~\ref{fig9} displays the errors in output weights updated by the projection algorithm and the weights trained offline for the two industry cases. The convergence of the output weights is evident, helping to prevent oscillatory or unstable behavior in the model when applied in practical industrial scenarios. This property is crucial for guaranteeing the smooth operation of systems, especially in industrial environments that require high precision and reliability. The introduction of the projection algorithm facilitates a rapid response to dynamic changes within the system, reduces computational overhead during the training process, and maintains robust performance when dealing with complex data. These findings validate the effectiveness of the proposed method and underscore its significant potential for modelling intricate industrial systems.

\section{Conclusion}
In this paper, a novel randomized learner model termed RSCN is presented for temporal data analytics. Distinguished from the well-known ESN, RSCN not only maintains the echo state property but also addresses problems such as the sensitivity of random parameter scales and manual structure setting. From the implementation perspective, firstly, the random parameters of the proposed approach are assigned in the light of a supervisory mechanism, and a special structure for the random feedback matrix is defined. Then, an online update of the output weights is implemented to handle the unknown dynamics. This hybrid framework ensures the universal approximation property for offline learning. Moreover, we also discuss the convergence of the output weights for online learning and the stability of the modelling performance. The effectiveness of the proposed method is verified by the MG time-series forecasting task, a nonlinear identification problem, and two real industrial datasets. The experimental results demonstrate that our proposed RSCNs outperform other models in terms of compact reservoir topology, as well as in learning and generalization performance.

It will be interesting to see more real industrial applications of the proposed schemes. Future research could investigate other methods to enhance the efficiency of RSCNs, such as alternative supervisory mechanisms or node selection strategies to accelerate error convergence. Furthermore, the proposed RSCN framework could be extended to other advanced randomized learner model \cite{ref39} to further enhance the model's learning, representation and explanation capabilities.

\end{CJK}

\begin{thebibliography}{99}
\bibitem{ref1}Q. Zhao, B. Yan, J. Yang and Y. Shi, “Evolutionary robust clustering over time for temporal data," IEEE Trans. Cybern., vol. 53, no. 7, pp. 4334-4346, Jul. 2023.

\bibitem{ref3}X. Sun, K. Liu, C. Wen, and W. Wang, “Predictive control of nonlinear continuous networked control systems with large time-varying transmission delays and transmission protocols," Automatica, vol. 64, pp. 76–85, Feb. 2016.

\bibitem{ref4}B. Ghojogh and A. Ghodsi, “Recurrent neural networks and long short-term memory networks: Tutorial and survey," arxiv:2304.11461, 2023.

\bibitem{ref5}X. Gao, L. Yan, G. Wang and C. Gerada, “Hybrid recurrent neural network architecture-based intention recognition for human–robot collaboration," IEEE Trans. Cybern., vol. 53, no. 3, pp. 1578-1586, Mar. 2023.

\bibitem{ref6}S. Hochreiter and J. Schmidhuber, “Long short-term memory," Neural Comput., vol. 9, no. 8, pp. 1735-1780, Nov. 1997.

\bibitem{ref7}A. Bala, I. Ismail, R. Ibrahim, and S. M. Sait, “Applications of metaheuristics in reservoir computing techniques: A review,” IEEE Access, vol. 6, pp. 58012-58029, 2018.

\bibitem{ref8}D. J. Gauthier, E. Bollt, A. Griffith, and W. A. Barbosa, “Next generation reservoir computing,” Nat. Commun., vol. 12, no. 1, pp. 1-8, Sep. 2021.

\bibitem{ref9}S. Scardapane and D. Wang, “Randomness in neural networks: An overview,” Wiley Interdiscip. Rev. : Data Min. Knowl. Discovery, vol. 7, no. 2, e1200, Feb. 2017.

\bibitem{ref10}H. Jaeger, “The echo state approach to analysing and training recurrent
neural networks with an erratum note,” in German National Research Center for Information Technology GMD Technical Report, Bonn, Germany, 2001.

\bibitem{ref11}C. Sun, M. Song, D. Cai, B. Zhang, S. Hong, and H. Li, “A systematic review of echo state networks from design to application,” IEEE Trans. Artif. Intell., vol. 5, no. 1, pp. 23-37, Jan. 2024.

\bibitem{ref12}Y. Han, Y. Jing, G. Dimirovski, and L. Zhang, “Multi-step network traffic prediction using echo state network with a selective error compensation strategy,” Trans. Inst. Meas. Control, vol. 44, no. 8, pp. 1656-1668, Oct. 2022.

\bibitem{ref13}Q. Chen, X. Li, A. Zhang and Y. Song, “Neuroadaptive tracking control of affine nonlinear systems using echo state networks embedded with multiclustered structure and intrinsic plasticity," IEEE Trans. Cybern., vol. 54, no. 2, pp. 1133-1142, Feb. 2024.

\bibitem{ref14}X. Chen, X. Luo, L. Jin, S. Li and M. Liu, “Growing echo state network with an inverse-free weight update strategy," IEEE Trans. Cybern., vol. 53, no. 2, pp. 753-764, Feb. 2023.

\bibitem{ref15}C. H. Valencia, M. Vellasco, and K. Figueiredo, “Echo state networks: novel reservoir selection and hyperparameter optimization model for time series forecasting,” Neurocomputing, vol. 545, no. 126317, Aug. 2023.

\bibitem{ref16}Y. Li and F. Li, “PSO-based growing echo state network,” Appl. Soft Comput., vol. 85, no. 105774, Dec. 2019.

\bibitem{ref17}J. Viehweg, K. Worthmann, and P. Mäder, “Parameterizing echo state networks for multi-step time series prediction,” Neurocomputing, vol. 522, pp. 214-228, Feb. 2023.

\bibitem{ref18}R. Soltani, E. Benmohamed, and H. Ltifi, “Echo state network optimization: A systematic literature review,” Neural Process. Lett., vol. 55, no. 8, pp. 10251-10285, Jun. 2023.

\bibitem{ref19}A. Rodan and P. Tino, “Minimum complexity echo state network," IEEE Trans. Neural Networks, vol. 22, no. 1, pp. 131-144, Jan. 2011.

\bibitem{ref20}J. Qiao, F. Li, H. Han, and W. Li, “Growing echo-state network with multiple subreservoirs," IEEE Trans. Neural Networks Learn. Syst., vol. 28, no. 2, pp. 391-404, Feb. 2017.

\bibitem{ref21}H. Jaeger, M. Lukoševičius, D. Popovici, and U. Siewert, “Optimization and applications of echo state networks with leaky-integrator neurons,” Neural Networks, vol. 20, no. 3, pp. 35-352, Apr. 2007.

\bibitem{ref22}C. Gallicchio and A. Micheli, “Deep echo state network: A brief survey,” arxiv:1712.04323, 2017.

\bibitem{ref23}D. Wang and M. Li, “Stochastic configuration networks: Fundamentals and algorithms,” IEEE Trans. Cybern., vol. 47, no. 10, pp. 3466-3479, Oct. 2017.

\bibitem{ref231}D. S. Broomhead and D. Lowe, “Radial basis functions, multi-variable functional interpolation and adaptive networks,” DTIC Document, vol. 28, Mar. 1988.

\bibitem{ref232}Y.-H. Pao and Y. Takefuji, “Functional-link net computing: Theory,
system architecture, and functionalities,” Computer, vol. 25, no. 5, pp. 76–79, May 1992.

\bibitem{ref24}H. A. B. Braake and G. V. Straten, “Random activation weight neural net (RAWN) for fast non-iterative training,” Engng Applic. Artif. Inter. vol. 8, no. 1, pp. 71-80, 1995.

\bibitem{ref25}B. Widrow, A. Greenblatt, Y. Kim, and D. Park, “The No-Prop algorithm: A new learning algorithm for multilayer neural networks,” Neural Networks, vol. 37, pp. 182-188, Jan. 2013.

\bibitem{ref35}G. Goodwin and K. Sin, “Adaptive filtering prediction and control,” Courier Corporation, 2014.

\bibitem{ref26}S. Billings, M. Korenberg, and S. Chen, “Identification of nonlinear output-affine systems using an orthogonal least squares algorithm,” Int. J. Syst. Sci., vol. 19, no. 8, pp. 1559–1568, Apr. 1988.

\bibitem{ref27}R. Kamalapurkar, N. Fischer, S. Obuz and W. E. Dixon, “Time-varying input and state delay compensation for uncertain nonlinear systems," IEEE Trans. Autom. Control, vol. 61, no. 3, pp. 834-839, Mar. 2016.

\bibitem{ref28}W. Li, Z. Zhang and S. S. Ge, “Dynamic gain reduced-order observer-based global adaptive neural-network tracking control for nonlinear time-delay systems," IEEE Trans. Cybern., vol. 53, no. 11, pp. 7105-7114, Nov. 2023.

\bibitem{ref29}G. Liu, J. H. Park, H. Xu and C. Hua, “Reduced-order observer-based output-feedback tracking control for nonlinear time-delay systems with global prescribed performance," IEEE Trans. Cybern., vol. 53, no. 9, pp. 5560-5571, Sept. 2023.


\bibitem{ref30}Y. Ebato, S. Nobukawa, and Y. Sakemi, “Impact of time-history terms on reservoir dynamics and prediction accuracy in echo state networks,” Sci. Rep., vol. 14, no. 1, Apr. 2024.

\bibitem{ref31}X. Wang, Y. Jin, W. Du, and J. Wang, “Evolving dual-threshold bienenstock-cooper-munro learning rules in echo state networks,” IEEE Trans. Neural Networks Learn. Syst., vol. 35, no. 2, pp. 1572-1583, Feb. 2024.

\bibitem{ref32}L. Cerina, M. Santambrogio, G. Franco, C. Gallicchio, and A. Micheli, “Echobay: design and optimization of echo state networks under memory and time constraints,” ACM Trans. Archit. Code Optim., vol. 17, no. 3, pp. 1-24, Aug. 2020.

\bibitem{ref33}Z. Carmichael, H. Syed, S. Burtner, and D. Kudithipudi, “Mod-deepesn: modular deep echo state network,” arxiv:1808.00523, 2018.

\bibitem{ref34}C. Yang, J. Qiao, H. Han, and L. Wang, “Design of polynomial echo state networks for time series prediction,” Neurocomputing, vol. 290, pp. 148-160, May 2018.

\bibitem{ref351}P. Lancaster and M. Tismenetsky, “The theory of matrices: with applications,” 2nd ed. San Diego, CA, USA: Academic Press, 1985.

\bibitem{ref36}L. Fortuna, S. Graziani, and M. G. Xibilia, “Soft sensors for product quality monitoring in debutanizer distillation columns,” Control Eng. Pract., vol. 13, no. 4, pp. 499-508, Apr. 2005.

\bibitem{ref39}D. Wang and M. J. Felicetti, “Stochastic configuration machines for industrial artificial intelligence,” arxiv:2308.13570v6, Oct. 2023.

\end{thebibliography}
\end{document}